\documentclass[journal,twoside,web]{ieeecolor}
\usepackage{tmi}
\usepackage{cite}
\usepackage{amsmath,amssymb,amsfonts}
\usepackage{algorithmic}
\usepackage{graphicx}
\usepackage{textcomp}
\usepackage{chapterbib}
\usepackage{docmute}
\usepackage{adjustbox}
\usepackage{amsmath}
\usepackage{bm}
\usepackage{arydshln}
\usepackage{multirow}
\usepackage{etoolbox}
\usepackage{xcolor}
\colorlet{blue}{black}
\newcommand{\todolist}[1]{{{#1}}}

\newcommand{\blue}[1]{\todolist{\textcolor{blue}{#1}}}
\makeatletter
\@ifundefined{color@begingroup}%
  {\let\color@begingroup\relax
   \let\color@endgroup\relax}{}%
\def\fix@ieeecolor@hbox#1{%
  \hbox{\color@begingroup#1\color@endgroup}}
\patchcmd\@makecaption{\hbox}{\fix@ieeecolor@hbox}{}{\FAILED}
\patchcmd\@makecaption{\hbox}{\fix@ieeecolor@hbox}{}{\FAILED}
\def\BibTeX{{\rm B\kern-.05em{\sc i\kern-.025em b}\kern-.08em
    T\kern-.1667em\lower.7ex\hbox{E}\kern-.125emX}}
\usepackage[switch]{lineno}

\begin{document}
% \title{A Co-evolutionary Framework: Towards Bridging Abnormality Detection And Report Generation}
\title{Unlocking the Potential of Weakly Labeled Data: A Co-Evolutionary Learning Framework for Abnormality Detection and Report Generation}
\author{Jinghan Sun, Dong Wei, Zhe Xu, Donghuan Lu, Hong Liu, Hong Wang, Sotirios A. Tsaftaris, \\ Steven McDonagh, Yefeng Zheng, \IEEEmembership{Fellow, IEEE} and Liansheng Wang
\thanks{Manuscript received \today.
This work was supported by National Natural Science Foundation of China (Grant No. 62371409) and Fujian Provincial Natural Science Foundation of China (Grant No. 2023J01005).
\textit{(J. Sun and D. Wei contributed equally to this work.)}
\textit{(Corresponding authors: L. Wang and Y. Zheng.)}}
\thanks{J. Sun, H. Liu, and L. Wang are with National Institute for Data Science in Health and Medicine, Xiamen University, Xiamen, China (email: jhsun@stu.xmu.edu.cn, liuhong@stu.xmu.edu.cn, lswang@xmu.edu.cn).}
\thanks{D. Wei, D. Lu, H. Wang, and Y. Zheng are with Jarvis Research Center, Tencent YouTu Lab (email: \{donwei, caleblu, hazelhwang, yefengzheng\}@tencent.com).}
\thanks{Y. Zheng is also with Medical Artificial Intelligence Laboratory, Westlake University, Hangzhou, China. (email: zhengyefeng@westlake.edu.cn)}
\thanks{Z. Xu is with the Department of Biomedical Engineering, The Chinese University of Hong Kong, Hong Kong, China (email: jackxz@link.cuhk.edu.hk).}
\thanks{Sotirios A. Tsaftaris and Steven McDonagh are with School of Engineering, The University of Edinburgh, Edinburgh, UK (email: \{s.tsaftaris, s.mcdonagh\}@ed.ac.uk)}
\thanks{J. Sun, H. Liu, and Z. Xu contributed to this work as interns at Jarvis Research Center, Tencent YouTu Lab.}
}

\maketitle

% \linenumbers

\begin{abstract}
Anatomical abnormality detection and report generation of chest X-ray (CXR) are two essential tasks in clinical practice. 
The former aims at localizing and characterizing cardiopulmonary radiological findings in CXRs, while the latter summarizes the findings in a detailed report for further diagnosis and treatment.
Existing methods often focused on either task separately, ignoring their correlation.
This work proposes a \textit{co-evolutionary} abnormality \textit{detection} and report \textit{generation} (CoE-DG) framework.
The framework utilizes both fully labeled (with bounding box annotations and clinical reports) and weakly labeled (with reports only) data to achieve mutual promotion between the abnormality detection and report generation tasks.
Specifically, we introduce a bi-directional information interaction strategy with generator-guided information propagation (GIP) and detector-guided information propagation (DIP). 
For semi-supervised abnormality detection, GIP takes the informative feature extracted by the generator as an auxiliary input to the detector and uses the generator's prediction to refine the detector's pseudo labels.
{\color{blue}We further propose an intra-image-modal self-adaptive non-maximum suppression module (SA-NMS).
This module dynamically rectifies pseudo detection labels generated by the teacher detection model with high-confidence predictions by the student.}
Inversely, for report generation, DIP takes the abnormalities' categories and locations predicted by the detector as input and guidance for the generator to improve the generated reports.
Finally, a co-evolutionary training strategy is implemented to iteratively conduct GIP and DIP and consistently improve both tasks' performance.
Experimental results on {\color{blue}two public CXR datasets} demonstrate CoE-DG's superior performance to several up-to-date object detection, report generation, and unified models. 
{\color{blue}Our code is available at https://github.com/jinghanSunn/CoE-DG.}

\end{abstract}

\begin{IEEEkeywords}
Abnormality detection, Report generation, Semi-supervised learning, Chest X-ray, Co-evolution
\end{IEEEkeywords}

\section{Introduction}
Chest X-ray (CXR) is the most commonly performed diagnostic radiograph in clinics, which helps spot abnormalities and diseases of the airways, blood vessels, heart, and lungs. 
This diagnostic workflow includes detecting anatomical abnormalities and writing a report to record the findings. 
Given the complexity and workload of clinical CXR reading, there is a growing interest in developing automated methods for anatomical abnormality detection \cite{qin2018computer} and radiology report generation \cite{monshi2020deep} in CXR---especially using deep neural networks (DNNs) \cite{lakhani2017deep,ougul2015lung,rajpurkar2017chexnet,liu2021exploring,chen2020generating}.
These methods are expected to expedite clinical workflow and reduce observational oversights.
Here, the detection task involves both localization (\textit{e.g.}, with bounding boxes) and characterization (\textit{e.g.}, cardiomegaly) of the abnormalities.
However, training accurate DNN-based detection models usually requires large-scale datasets with high-quality per-abnormality annotations, which is costly in time, effort, and expense.
Meanwhile, the visual bias problem, where normal regions dominate a CXR image over abnormal, poses a challenge for accurately spotting tiny abnormal regions by the report generation model. 

% (Weakly supervised detection?)

To relieve the annotation burden for abnormality detection, a few works \cite{chen2022label,sohn2020simple,xu2021end} proposed semi-supervised object detection and achieved noteworthy advances in the natural image domain. Most of these methods %\cite{tarvainen2017mean,xie2020self} 
were built on the teacher-student distillation (TSD) paradigm \cite{hinton2015distilling}.
In TSD, a teacher model is firstly trained on the labeled data. 
Then, a student model is trained on both the labeled data with real annotations and the unlabeled data with pseudo labels generated (predicted) by the teacher.
However, compared with objects in natural images, the abnormalities in CXR can be subtle and less well-defined with ambiguous boundaries, thus likely introducing great noise to the pseudo labels and eventually leading to suboptimal performance of semi-supervised learning with TSD.

To alleviate the visual bias problem in report generation, some approaches\cite{liu2021exploring,you2021aligntransformer} proposed aligning predefined disease tags and corresponding visual region features to make the networks focus on abnormal regions related to those tags.
%some approaches\cite{liu2021exploring,you2021aligntransformer} have been proposed, which adopted indirect methods (\textit{e.g.}, aligning the pre-defined disease tags and corresponding visual region features) trying to provide more information, making the network focus on the abnormal regions related to those tags.
However, they required domain expertise and extra labor to define and extract the disease tags, and it was still uncertain whether these models accurately identified abnormal regions while generating reports for input CXR images.
Furthermore, most report generation methods prioritized literal consistency but ignored clinical efficacy. 
Lastly, despite the encouraging performance of the above methods in addressing the detection and generation tasks separately, none of them attempted to bridge these two tasks.

We are aware of a series of works\cite{yang2021crossing,cho2021unifying,gupta2022towards,hu2021unit} aimed at obtaining a unified framework for various vision-language tasks (\textit{e.g.}, visual question answering and image captioning) in the natural image domain.
% Most of these methods pre-trained a unified model by learning a general representation that can be used for both language and vision tasks. 
% most of which adopted a shared feature extraction backbone with an individualized branch for each downstream task.
% However, the correlations between these tasks have not been fully utilized during training.
However, these methods required data with paired bounding box and caption annotations for training, thus could not utilize weakly labeled images (\textit{i.e.}, with captions but no bounding boxes).
% \textbf{However, they had to fine-tune for each vision and language task and were incapable of jointly optimizing for performance improvement.}

This work presents a co-evolutionary abnormality detection and report generation (CoE-DG) framework for bridging both tasks in CXR.
{\color{blue}Text reports describe significant findings in CXRs and are readily available for most archive radiographs.
They are a valuable source of image-level supervision signals unique to medical image data.}
Unlike most existing unified frameworks that overlooked valuable knowledge in the weakly labeled data, we design an iterative optimization method that fully utilizes them, enhancing the information interaction between vision and language tasks and promoting joint performance improvement. 
This approach includes a report generator guided information propagation (GIP) module for anatomical abnormality detection and an abnormality detector guided information propagation (DIP) module for report generation. 

Specifically, through GIP, we first enrich the detector's input features by incorporating the generator's informative features. 
Then, based on TSD \cite{hinton2015distilling}, the generator's auxiliary categorical prediction is used for noise reduction in the pseudo labels of the detection task. 
Besides the GIP, we additionally propose self-adaptive non-maximum suppression \cite{girshick2014rich} (SA-NMS) for intra-image-modal refinement in anatomical abnormality detection.
The predictions by both the teacher and student detectors go through NMS together to produce new pseudo detection labels for training.
In this way, the pseudo labels generated by the teacher detector are dynamically rectified by high-confidence predictions of the student who is getting better as training continues.
Reversely, through DIP, the abnormality tokens and location embeddings extracted from the detector's output serve as input to the generator. 
Moreover, abnormality categories predicted by the detector are used as pseudo labels to supervise the generator with an auxiliary multi-label prediction task, improving the fidelity of generated diagnostic reports.
Finally, GIP and DIP alternate in a loop, where either model is trained while fixing the other and using the other's prediction for performance boosting via information interaction.

In summary, our contributions are three-fold:
\begin{itemize}
  \item We propose a co-evolutionary framework, CoE-DG, that bridges the detector and generator, optimizing them iteratively to achieve mutual benefits to anatomical abnormality detection and radiology report generation in CXR.
  After training, the framework can simultaneously detect accurate abnormalities and generate high-quality radiology reports given a CXR.
  \item We propose GIP for semi-supervised anatomical abnormality detection.
  GIP incorporates extensive information from the generator to refine pseudo labels of the teacher detector for noise reduction. Additionally, we propose SA-NMS for dynamic intra-image-modal pseudo label refinement.
  \item We propose DIP for report generation. DIP uses the fine-grained information extracted by the detector and supervises the generator using pseudo labels predicted by the detector.
\end{itemize}
Thorough experiments on two public CXR datasets \cite{boecking2022making,tam2020weakly} derived from MIMIC-CXR \cite{johnsonmimic} demonstrate that CoE-DG effectively enhances both anatomical abnormality detection and report generation performance compared to previous state-of-the-art (SOTA) methods.

This work builds on our preliminary exploration \cite{sun2023you} yet is distinct from the earlier work and presents novel methodologies in three main aspects.
% This work is a comprehensive extension of our preliminary exploration \cite{sun2023you} in three main aspects.
Most prominently, we propose a new and improved framework for simultaneously detecting abnormal regions and generating a report for a given CXR.
The new framework has greater clinical significance than the earlier work, which could only detect abnormal regions.
Secondly, we further improve the anatomical abnormality detection performance by enriching the input feature and denoising the pseudo labels based on the report generation model.
Thirdly, we propose a novel report generation method that leverages abnormality tokens and location embeddings extracted from the output of the detection model as input.
The method also incorporates an auxiliary multi-label classification task to enhance the fidelity of generated reports.

{\color{blue}Table \ref{tab:list-of-abbr} provides a list of abbreviations used in this paper.}

\begin{table}[]\color{blue}
\setlength{\tabcolsep}{0.8mm}
\caption{Lists of abbreviations. 
The page on which each abbreviation is first used is listed, too.}
\label{tab:list-of-abbr}
\begin{adjustbox}{width=\linewidth}
\begin{tabular}{l|l|c}
\hline
Abbreviation & Definition    & First use                                              \\ \hline
CXR           & Chest X-ray     &  Page 1                                          \\ 
CoE-DG        & Co-evolutionary abnormality detection and & Page 1 \\ 
              & report generation & \\ 
GIP           & Generator-guided information propagation            & Page 1\\  %/Page 4
DIP           & Detector-guided information propagation        & Page 1\\  %/Page 6
SA-NMS        & Self-adaptive non-maximum suppression                       & Page 1\\
DNN          & Deep neural networks                               & Page 1\\ 
TSD           & Teacher-student distillation                                & Page 1\\  %/Page 4
LLM           & Large language model        & Page 3 \\  
LMM           & Large multimodal model        & Page 3 \\  

PD-CXR & Pneumo-disease chest X-ray & Page 7 \\
mAP & Mean average precision & Page 7\\
IoU & Intersection over union & Page 7 \\
AUC & Area under the curve & Page 7\\
SOTA & State-of-the-art    & Page 8 \\ 
MLMM & Medical large multimodal model & Page 8 \\
% CAM & Class activation map & Page 8 \\
TNR & True negative rate & Page 9 \\
EMA & Exponential moving average & Page 10\\
FE & Feature enhancement & Page 11\\
PLR & Pseudo labels refinement & Page 11 \\
AT & Abnormality token & Page 11\\
LE & Location embedding & Page 11 \\
Cls. Sup. & Classification supervision & Page 11\\
GFLOP & Giga floating-point operation & Page 11\\

 \hline
\end{tabular}
\end{adjustbox}
\end{table}

\section{Related Work}
\subsubsection{Object Detection}
Object detection is a fundamental task in computer vision that identifies and locates objects within images and videos.
There are two mainstream approaches: single-stage and two-stage. 
A single-stage object detector, such as YOLO\cite{redmon2016you} and RetinaNet\cite{lin2017focal}, directly predicts objects' bounding boxes and class probabilities in one shot. 
This approach is fast and suitable for real-time applications. 
A two-stage object detector, such as Faster R-CNN\cite{girshick2015fast} and Mask R-CNN\cite{he2017mask}, first proposes a set of candidate regions and then refines them to get the final bounding boxes and class probabilities. 
In this work, we use RetinaNet\cite{lin2017focal} as our detection framework due to its prominent performance in RSNA Pneumonia Detection Challenge\cite{Cheng2019}. 
However, existing fully supervised methods require a large amount of labeled data for training, making the annotation process time-consuming and expensive.

To reduce the labeling burden, some researchers proposed using self-supervised\cite{liu2020self} or weakly supervised\cite{zhang2021weakly,huang2020comprehensive} methods. 
Self-supervised detection \cite{liu2020self} typically learned from pretext tasks to enforce invariance across different augmented views of an image (such as cropped or rotated) without the need for labels in the training stage.
% , and labels are required during testing.}
Other works \cite{huang2020comprehensive,cheng2020high} resorted to learning object detectors with weak supervision that only needed accessible weak annotations, such as image-level labels. 
Most weakly supervised object detection methods, in the natural image domain, used the class activation map (CAM)~\cite{zhou2016learning} to generate a heatmap highlighting the regions of an image that contributed most to a particular class prediction. 
This heatmap was used to localize the object within the image. 
For medical image object detection, a few works \cite{bhalodia2021improving,tam2020weakly,yu2022anatomy} utilized radiology reports as a form of weak supervision for localizing pneumonia and pneumothorax in CXR.
However, studies have shown that there are still apparent gaps in performance between image-level self-/weakly- supervised and bounding-box-level fully supervised detection \cite{bearman2016s,ji2022point}.

Alternatively, seeking a trade-off between annotation effort and model performance, semi-supervised learning aims to achieve reasonable performance with an acceptable quantity of manual annotations.
Semi-supervised object detection methods have achieved noteworthy advances in the natural image domain, including two categories: consistency-based \cite{jeong2019consistency,tang2021proposal} and pseudo-label methods\cite{chen2022label,sohn2020simple,xu2021end}.
Consistency-based methods enforce consistency between object location predictions derived from both noisy and original proposals. 
In contrast, pseudo-label methods use a teacher-student distillation (TSD) paradigm to train models on labeled and unlabeled data.
A study \cite{xu2021end} showed that pseudo-label methods were better than consistency-based.
Despite its potential for reducing annotation burdens, the TSD paradigm faces challenges in detecting subtle abnormalities with unclear boundaries in medical images, such as CXR, leading to suboptimal performance.
To address this issue, we propose a pseudo label refinement method that utilizes the information from the report generation model to filter out noisy and unreliable pseudo labels, thereby improving the performance of the detection model.

\subsubsection{Radiology Report Generation}
Radiology report generation automatically generates reports summarizing the findings and impressions from medical images.
The encoder-decoder framework is commonly used in most medical report generation methods \cite{jing2017automatic,xue2018multimodal,yuan2019automatic}, which typically produces a single descriptive sentence. 
However, generating a comprehensive radiology report requires a long paragraph with multiple structured sentences, each describing a specific medical observation in a particular image region.
In addition, the dominance of normal regions in CXR poses a visual bias problem, making it challenging for report generation models to catch and describe small abnormal regions. 
Various approaches \cite{liu2021exploring,chen2020generating,you2021aligntransformer} have been proposed to overcome this bias, which altered the network's attention by aligning the disease tags and their corresponding visual region embeddings.
However, it is unclear if these methods accurately captured abnormal regions.
Furthermore, most report generation methods prioritized literal consistency over medical accuracy, which needs to be addressed for practical clinical application.
Alternatively, we propose utilizing fine-grained features extracted from a detection model to generate reports, providing the network with direct information about abnormal regions. 
In addition, we incorporate an auxiliary multi-label classification task to encourage the generation model to attend to abnormalities in CXRs besides mastering fluent language.
{\color{blue}Lastly, \cite{jing2017automatic} and \cite{jin2024promptmrg} aim to leverage the predictions of abnormalities to guide high-quality report generation a motivation shared with our approach.
Our framework takes a step further to enable simultaneous localization of various abnormalities in CXRs besides generating reports, thus providing extra clinical value.
}
%leveraging the predictions of abnormalities to guide the report generation

\subsubsection{Unified Framework}
Recently, researchers in the natural image domain introduced the concept of a unified model that simultaneously tackles vision and language tasks such as question answering, image classification, and text generation.
{\color{blue}CLIP \cite{radford2021learning} laid the foundation for many recent advances in cross-modal alignment, pretraining, and processing by large-scale vision-language contrasting.
For biomedical vision-language processing, BioMedCLIP \cite{zhang2023biomedclip} proposed domain-specific large-scale pretraining using 15 million figure-caption pairs extracted from biomedical research articles.
It achieved impressive performance on various benchmark biomedical imaging tasks.
}
Several studies suggested that combining multiple tasks during training was essential for achieving general intelligence \cite{hu2021unit,gupta2022towards}. 
By merging the losses of various tasks, it was possible to obtain a domain-agnostic model with shared parameters that could benefit downstream tasks.
While achieving decent results, these models often required large quantities of fine-grained annotations, such as box annotations and paired captions for each image. 
However, obtaining an extensive collection of annotations can be difficult in practice, especially for medical images.
To the best of our knowledge, in medical imaging, we are the first to propose a mutually reinforcing co-evolution strategy that fully utilizes weakly labeled data to enhance the performance of image-text tasks. 
This approach relieves the burden of obtaining large-scale fine-grained annotations.

% \subsubsection{Large Vision-Language Model}
% \todo{
% Large language models have become increasingly popular in recent years due to their impressive performance across a wide range of natural language processing (NLP) tasks. The Generative Pre-trained Transformer (GPT) \cite{radford2018improving}, an influential language model based on the Transformer architecture, is pre-trained using a unidirectional language modeling objective. This enables GPT to predict the next word in a sentence given the previous words. With iterative improvement, the latest version, GPT-4 \cite{achiam2023gpt}, has demonstrated remarkable performance on various NLP tasks. }

\begin{figure*}[t]
\centering
\includegraphics[width=0.95\textwidth]{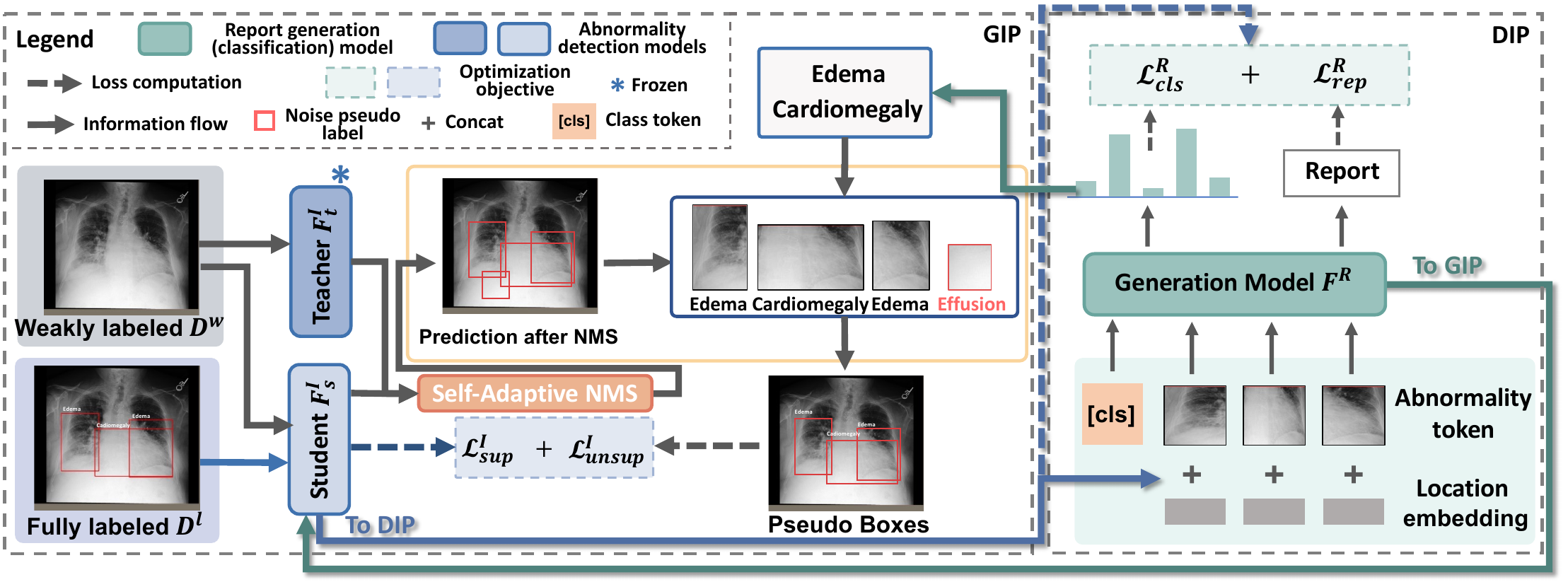}
\caption{Overview of the proposed framework.
{\color{blue}NMS: non-maximum suppression;}
GIP: generator-guided information propagation;
DIP: detector-guided information propagation.
{\color{blue}For semi-supervised abnormality detection, a self-adaptive NMS module dynamically rectifies pseudo detection labels generated by the teacher detection model $F^I_t$ with high-confidence predictions by the student $F^I_s$.
The GIP takes the feature extracted by the generator $F^R$ as an auxiliary input to $F^I_s$;
it also uses $F^R$’s prediction to refine the pseudo labels further.
Inversely, for report generation, the DIP takes the abnormalities detected by $F^I_s$ as input and guidance to $F^R$ to improve generated reports.
The abnormalities' categories predicted by $F^I_s$ are also used to supervise $F^R$'s training via $\mathcal{L}^R_{cls}$ for weakly labeled samples.
}
% \sout{cls: class token.
% The framework employs a pretrained detection model $F_t^I$. The generator-guided information propagation (GIP) refines pseudo-detection labels and augments detection features, improving abnormality localization. Concurrently, the detector-guided information propagation (DIP) uses inputs from the detection model to enhance report generation. These two processes co-evolve during training, optimizing both the detection model $F_t^I$ and the report generation model $F^R$ for improved performance in CXR analysis.}
}
\label{fig:procedure}
\end{figure*}

% \todo{The integration of visual information into LLMs has given rise to multimodal models capable of effectively processing and generating content based on both textual and visual inputs.
% One notable example is the Visual-Linguistic BERT (ViLBERT) \cite{lu2019vilbert}, which extends the BERT architecture to incorporate visual information and textual data. 
% However, applying general-purpose LLMs or multimodal LLMs to the medical domain can be challenging, as medical texts often contain specialized terminology and structured syntax that may not be adequately represented in the training data.
% To this end, MedPaLM \cite{singhal2022large} proposed a new instruction prompt tuning method for adopting LLMs into the medical domain. Furthermore, LLaVA-Med aimed to extend multi-modal instruction tuning to create a multimodal conversational assistant.
% These methods use image-caption pairs from a wide spectrum of medical literature.
% Despite these advances, few works have focused on X-ray report generation or abnormality detection.
% Therefore, in this paper, we design a specialized model for report generation and abnormalities detection for more accurate diagnosis rather than serving general purposes.
% }
\subsubsection{Large Language Models}
{\color{blue}Recently, large language models (LLMs) such as ChatGPT \cite{achiam2023gpt} and LlaMA \cite{touvron2023llama} have gained widespread and growing interest due to their unprecedented capabilities in language understanding and generation. 
Research efforts have been devoted to investigating their applications in various domains, including medicine \cite{gu2021domain,lee2020biobert}. In addition, integrating visual information into LLMs has given rise to large multimodal models (LMMs) capable of understanding both textual and visual inputs \cite{alayrac2022flamingo,wang2023cogvlm,liu2024visual}. However, directly applying general-purpose LMMs to medical images would yield unsatisfactory outcomes, mainly due to 1) the domain gaps between natural and medical images and 2) the specialized medical terminology. 
To this end, Li et al. \cite{li2024llava} presented a Large Language and Vision Assistant for BioMedicine (LLaVA-Med). 
The key idea was to leverage a large-scale, broad-coverage biomedical figure-caption dataset extracted from PubMed Central, use GPT-4 to self-instruct open-ended instruction-following data from the captions and fine-tune a general-domain LMM.
Yet LLaVA-Med was targeted as an instruction-following conversational assistant for versatile biomedical images, thus falling short in competency for CXR reporting in comparison with leading specialized models. 
On the contrary, XrayGPT \cite{thawkar2023xraygpt} and CheXagent \cite{chen2024chexagent} were LMMs specifically proposed for CXR interpretation, demonstrating competitive performance on the task of report generation. 
In addition to generating text reports our framework can accurately localize the abnormalities in CXRs, given a relatively small quantity of annotations for training, which is currently challenging for the considered LMMs.}

\section{Methods}
\subsection{Problem Setting}
{\color{blue}We consider a training dataset consisting of a fully labeled subset $D^l=\{(x,\{r_1, \ldots, r_T\},A)\}$ and a weakly labeled subset $D^w=\{(x,\{r_1, \ldots, r_T\})\}$, where $x$ is a CXR, $\{r_1, \ldots, r_T\}$ is the accompanying report with $T$ tokens, $r_t\in \mathbb{V}$, $\mathbb{V}$ is the vocabulary of all possible tokens, $A = \{(y^l, B^l)\}$ is the abnormality annotation for a fully labeled sample including both bounding boxes (\textit{i.e.}, location) $\{B^l\}$ and corresponding categories (\textit{i.e.}, classification) $\{y^l\}$.}
For practical usability, we have $|D^l| \ll |D^w|$ to significantly reduce the burden of manually annotating abnormalities.
%The training data set for semi-supervised anatomical abnormality localization and fully supervised report generation consists of both weakly labeled samples $D_u=\{(x_i^u,\{r_1, \ldots, r_T\}_i^u)\}_{i=1}^{N_u}$ and \textcolor{red}{fully} labeled ones $D^l=\{(x_i^l,\{r_1, \ldots, r_T\}_i^l,A_i)\}_{i=1}^{N_l}$, where $x$ and $\{r_1, \ldots, r_T\}, r_t\in \mathbb{V}$ are a CXR and accompanying report, respectively, $T$ is the length of reports, $\mathbb{V}$ is a vocabulary of all possible tokens, $A_i = \{(y^l, B^l)\}$ is the annotation for a fully labeled sample including both bounding boxes $\{B^l\}$ and corresponding categories $\{y^l\}$, and $N^l\ll N^u$ for practical use scenario. 
The objective is to obtain from $D^l \cup D^w$ a detector that can accurately localize and correctly identify the abnormalities, and a generator that can produce a correct textual description report in any test CXR.
%This will be achieved by utilizing both the \textcolor{red}{fully} labeled and weakly labeled CXRs in the training set and encouraging interaction between the detection and generation model.

\subsection{Method Overview}
An overview of our framework is shown in Fig. \ref{fig:procedure}.
Suppose a pretrained detection model $F^I_{t}$ (\textit{e.g.}, on fully labeled data) for abnormality detection in CXR is given. 
We employ a bidirectional co-evolutionary method for information interaction. 
One direction is conducted by generator-guided information propagation (GIP). GIP takes the informative feature from the generator as an auxiliary input to the student detector and uses the generator's predictions to refine the pseudo detection labels of the teacher detection model.
% and enhance the detection feature. 
Meanwhile, self-adaptive non-maximum suppression (SA-NMS) also filters the pseudo labels.
By this, we obtain refined pseudo labels to supervise the student detection model $F_s^I$ and informative detection features toward better anatomical abnormality localization. 
Another direction is conducted by detector-guided information propagation (DIP).
DIP passes the abnormality token, location embedding, and pseudo labels generated by the detector $F_s^I$ to the report generator $F^R$.
This process helps improve the report generation performance with abnormality-guided multi-label classification and generation feature enhancement. 
Reversely, the improved generation model $F^R$ helps train a better detection model $F^I_s$ via GIP. 
Thus, both types of models co-evolve during training.
After training, we use the student detection model $F_s^I$ for abnormality localization and the generation model $F^R$ for report generation in test CXRs.

\subsection{Preliminary Pseudo Label Distillation for Semi-supervised Detection}
Our baseline semi-supervised detection model follows the teacher-student knowledge distillation (TSD) procedure \cite{hinton2015distilling}.
A student detection model $F_s^I$ is trained in the semi-supervised setting by distilling from a teacher detection model $F_t^I$ trained on labeled CXRs, with the loss function:
\begin{equation}
\label{eq:pld}
\begin{aligned}
    \mathcal{L}^{I} = \mathcal{L}_\mathrm{sup}^{I} &+ \mathcal{L}_\mathrm{unsup}^{I}
    ={\sum}_{D^l}\big[\mathcal{L}_\mathrm{cls}^{I}\left({\hat{y}}, {y^l}\right)+ \mathcal{L}_\mathrm{reg}^{I}\big({\hat{B}}, {B^l}\big)\big]\\
&+{\sum}_{D^w}\big[\mathcal{L}_\mathrm{cls}^{I}\left({\hat{y}}, {y^t}\right)+ \mathcal{L}_\mathrm{reg}^{I}\big({\hat{B}}, {B^t}\big)\big],
\end{aligned}
\end{equation}
where $\{(\hat{y},\hat{B})\}=F^I_s(x)$ are predictions by the student model, $F^I_s=f^I_{det}(f^I_{feat})$ consists of a feature extraction backbone network $f^I_{feat}$ and a detection network $f^I_{det}$,  $\{(y^t,B^t)\}=F_t^I(x)$ are pseudo labels generated by the teacher model (note that for reliability, we only retain pseudo labels with prediction possibilities above a threshold, which is set to 0.9 according to Table \ref{tab:ablation_threshold}), %where $\{(\hat{y},\hat{B})\}=\{f_{s,cls}(e^I),f_{s,box}(e^I)\}=F^I_s(x)$ are the predictions by the student model, $F^I_s$ is formed by a feature extractor $f_{s,ext}$, a class subnet $f_{s,cls}$ and a box subnet $f_{s,box}$, $e^I = f_{s,ext}(x)$ is the feature extracted by a Feature Pyramid Network (FPN) backbone $f_{s,ext}$ on top of a feedforward ResNet architecture, $\{(y^t,B^t)\}=F_t^I(x)$ are the pseudo class and bounding box labels generated by the teacher model
$\mathcal{L}_\mathrm{cls}^{I}$ is the focal loss \cite{lin2017focal} for classification, and $\mathcal{L}_\mathrm{reg}^{I}$ is the smooth L1 loss \cite{lin2017focal} for bounding box regression.
In each batch, fully and weakly labeled instances are sampled according to a controlled ratio.
The resulting detection model $F_s^I$ will be utilized later to help radiology report generation.

\subsection{Self-Adaptive Non-maximum Suppression}
During the TSD, the teacher detection model $F_t^I$ is frozen.
While its knowledge suffices for guiding the student detection model $F_s^I$ in the early stage of TSD, it may somehow impede the learning of $F_s^I$ when $F_s^I$ gradually improves by also learning from a large amount of weakly labeled data.
Therefore, to gradually improve the quality and robustness of the pseudo detection labels as $F_s^I$ learns,
we propose to perform self-adaptive non-maximum suppression \cite{girshick2014rich} (SA-NMS) to combine the pseudo labels $\{(y^t,B^t)\}$ output by $F_t^I$ and the predictions $\{(\hat{y},\hat{B})\}$ by $F_s^I$ in each mini-batch.
Specifically, we perform NMS on the combined set of the pseudo labels and predictions: 
\begin{equation}\label{eq:NMS}
\{(y^\mathrm{NMS},B^\mathrm{NMS})\}= \mathrm{NMS}\big(\{(y^t,B^t)\} \cup \{(\hat{y}, \hat{B})\}\big),     
\end{equation}
and replace $\{(y^t,B^t)\}$ in Eqn.  (\ref{eq:pld}) with $\{(y^\mathrm{NMS},B^\mathrm{NMS})\}$ for supervision by weakly labeled CXRs.
In this way, highly confident predictions by the maturing student can rectify imprecise ones by the teacher, leading to better supervision signals stemming from \textcolor{blue}{weakly labeled data}.

% \subsection{Preliminary Report Generation?}

\begin{figure*}[htb]
\centering
\includegraphics[width=0.9\textwidth]{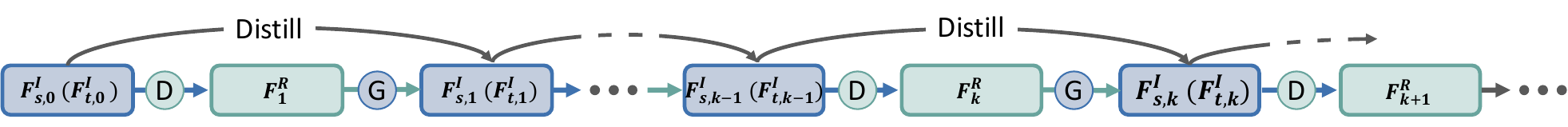}
\caption{Illustration of the co-evolution strategy.
``D'' and ``G'' represent detector-guided and generator-guided information propagation (DIP and GIP), respectively.
{\color{blue}The $k$\textsuperscript{th} iteration student detection model $F_{s,k}^I$ is distilled from the teacher $F_{t,k-1}^I$ guided by the generation model $F_{k}^R$ via GIP. 
Subsequently, $F_{s,k}^I$ is frozen and used to 1) guide the training of the $(k+1)$\textsuperscript{th} generation model $F_{k+1}^R$ via DIP, and 2) serve as the teacher detection model in the next iteration, \textit{i.e.}, $F_{s,k}^I\rightarrow F_{t, k}^I$.}}
\label{fig:co-evolution}
\end{figure*}

\subsection{Co-evolutionary Information Interaction}
{
% Although significant progress has been made in object detection\cite{sohn2020simple,xu2021end} and report generation\cite{xue2018multimodal,yuan2019automatic} separately, 
Some studies\cite{hu2021unit,gupta2022towards} explored the development of a unified network, encouraging learning vision-language cross-task knowledge to tackle multiple tasks.
However, these studies failed to utilize the treasured knowledge in weakly labeled data.}
This paper proposes a learning framework based on a co-evolutionary information interaction strategy, including generator-guided information propagation (GIP) and detector-guided information propagation (DIP). This strategy effectively leverages weakly labeled data to connect anatomical abnormality detection and report generation in a cycle.

\subsubsection{Detector-guided Information Propagation (DIP)}
Finding small abnormal regions and accurately generating disease-describing text in CXR is typically challenging for the report generation model.
Meanwhile, the detection model is expected to localize abnormalities accurately.
Therefore, we propose detector-guided information propagation (DIP) to help train the report generation model.
Concretely, given a CXR image $x$, we obtain a set of abnormalities $\{(y^I,B^I)\}$.
For a fully labeled sample, \textit{i.e.}, $x \in D^l$, we directly use the manual annotation $\{(y^l,B^l)\}$ as $\{(y^I,B^I)\}$.
For a weakly labeled sample $x \in D^w$, $\{(y^I,B^I)\}$ consists of the set of abnormalities $\{(\hat{y},\hat{B})\}$ predicted by the student detection model $F_s^I$ with prediction possibilities above 0.9.
Then, we extract an abnormality token $e^{R}$ for each abnormality in $\{(y^I,B^I)\}$ by feeding the corresponding image patch cropped according to $B^I$ through a projection network $f^R_{prj}$.
Suppose there are less than five abnormalities (the maximum number of abnormality bounding boxes for a CXR in the training set). 
In that case, we use special NULL tokens to make up the shortage for a consistent token length of input images.
Meanwhile, the bounding box is scaled to percentages with respect to the width and height of the image: $\lfloor (b_0/W, b_1/H, b_2/W, b_3/H) * 100\rfloor$, where $(b_0, b_1)$ and $(b_2, b_3)$ are coordinates of the top-left and bottom-right corners of the box, $W$ and $H$ are width and height of the images, respectively, and $\lfloor \cdot \rfloor$ indicates flooring. 
It is then projected to a location embedding $e^{loc}$ of the same dimension as the abnormality token.
Next, $e^R$ and $e^{loc}$ are concatenated to represent the bounding box.
In addition, motivated by ViT \cite{dosovitskiy2020image}, we prepend a class token $e^{cls}$ to aggregate information from all abnormality tokens and apply a classification head to predict $\{y^I\}$.
Thus, we have
\begin{equation}\label{eq:gen}
    \{\hat{y}^R\}, \{\hat{r}_{1}, \ldots, \hat{r}_{T}\}=F^{R}\left({e^{cls}}, \{[{e^{R}}, {e^{loc}}]\}\right),
\end{equation}
where $\{\hat{y}^R\}$ are the abnormality categories predicted on the class token, $\{\hat{r}_{1}, \ldots, \hat{r}_{T}\}$ compose the generated report, and $[\cdot]$ indicates concatenation.
{\color{blue}Here, $F^R$ can be any typical report generation model with the generalized form of an encoder followed by a decoder $f^R_{dec}(f^R_{enc}(\cdot))$, such as R2Gen \cite{chen2020generating} and R2GenCMN \cite{chen2022cross}, plus a classification head $f_{cls}^R$.
% $F^R(\cdot)=f^R_{dec}(f^R_{enc}(\cdot))$ is composed of a variant of the Transformer \cite{vaswani2017attention} encoder $f^R_{enc}$ and decoder $f^R_{dec}$.
First, the class token, abnormality tokens, and location embeddings are encoded into hidden states:
\begin{equation}
    h^{cls}, \{h^{R}\} = f_{enc}^R\big({e^{cls}}, \{[{e^{R}}, {e^{loc}}]\}\big),
\end{equation}
where $h^{cls}$ and $\{h^{R}\}$ represent the hidden states of the class token and the detected abnormalities, respectively.
Then, $h^{cls}$ and $\{h^{R}\}$ are fed into $f_{cls}^R$ and $f^R_{dec}$, respectively:
\begin{equation}
   \{\hat{y}^R\} =  f^R_{cls}(h^{cls}),\ \hat{r}_t = f^R_{dec}\big(\{h^{R}\}, \hat{r}_1, \ldots, \hat{r}_{t-1}\big).
\end{equation}
% \begin{equation}
%    \{\hat{y}^R\} =  f^R_{cls}(\{h^{cls}\}), \{\hat{r}\} = f^R_{dec}(\{h^{R}\}, \hat{r}_1, \ldots, \hat{r}_{t-1}),
% \end{equation}
Thus, the entire generation process can be formalized as a recursive application of the chain rule:
\begin{equation}
    p\big(\{\hat{r}_{1}, \ldots, \hat{r}_{T}\}\big)=\prod_{t=1}^{T}p\big(\hat{r}_t|\{h^R\},\hat{r}_{1}, \ldots, \hat{r}_{t-1}\big).
\end{equation}
% where the conditional probability of the next token ($\hat{y}^R_t$) given the input sequence ($h^{R}$) and the previously generated tokens is calculated as follows:
% \begin{equation}
%     p(\hat{r}_{T} | \{h^{R}\}, \hat{r}_1, \ldots, \hat{r}_{t-1}) = \text{softmax}(f^R_{dec}(\{h^{R}\}, \hat{r}_1, \ldots, \hat{r}_{t-1}))
% \end{equation}
Finally, supervised by $y^I$ (either ground truth or pseudo labels) and the ground truth report $\{r_1, \ldots, r_T\}$, the generator $F^R$ is trained by
\begin{equation}\label{eq:rgen}
\begin{aligned}
\mathcal{L}^R=&\mathcal{L}_\mathrm{cls}^{R}+\mathcal{L}^R_\mathrm{rep}={\sum}_{D^l \cup D^w}\mathcal{L}_\mathrm{CE}(y^I, \hat{y}^R)\\
&-{\sum}_{D^l \cup D^w}\log p\big(\{r_1, \ldots, r_T\}\big),
\end{aligned}
% \begin{aligned}
% \mathcal{L}^R=\mathcal{L}_\mathrm{cls}^{R}&+\mathcal{L}^R_\mathrm{rep}={\sum}_{D^l \cup D^w}\mathcal{L}_\mathrm{CE}(y^I, \hat{y}^R)\\ -&{\sum}_{D^l \cup D^w}\log p\left(\hat{r}_{T} \mid \hat{r}_1, \ldots, \hat{r}_{T-1}; \theta^{R} \right),
% \end{aligned}
\end{equation}
where $\mathcal{L}_\mathrm{CE}$ is the cross-entropy loss.}
% and $\theta^{R}$ is the parameters of the generation model $F^R$.

% Doing so 
Feeding the abnormality detection results to the generator provides precise information about the abnormalities' locations and types to enhance its understanding of the CXR.
Furthermore, incorporating the auxiliary multi-label classification task enhances the generator's sensitivity to catch and describe diseases.

\subsubsection{Generator-guided Information Propagation (GIP)}
While training the report generation model, the class token aggregates information from all abnormality tokens. 
% and theoretically contains more information
Additionally, the cross-attention mechanism in the generator's architecture between various abnormality tokens results in a comprehensive understanding of the CXR image that is different from the detector.
We thus expect that using the class token's classification prediction to filter pseudo-labels of the detection model can improve the detection performance.
Furthermore, we propagate the image embeddings extracted by the generator's projection network $f^R_{prj}$ to the detector for feature enhancement.
% Hence, we can propagate this enhanced feature to the detection model for its benefit.
Concretely, given a weakly labeled image $x\in D^l$, we feed it to the student detector's feature extraction network to obtain an image embedding $e^I=f^I_{feat}(x)$ and to $f^R_{prj}(x)$ to obtain another embedding $e^{IR}=f^R_{prj}(x)$.
These two embeddings are concatenated to form an enhanced image embedding, which is fed to the student detection network to yield $\{(\hat{y},\hat{B})\}=f^I_{det}([e^I, e^{IR}])$.
Then, we get $\{(y^\mathrm{NMS},B^\mathrm{NMS})\}$ from $\{(\hat{y},\hat{B})\}$ by the proposed SA-NMS (Eqn. (\ref{eq:NMS})).
Meanwhile, we also obtain the set of abnormality categories $\{\hat{y}^R\}$ predicted by the generator on the class token $e^{cls}$ (Eqn. (\ref{eq:gen})).
% and a set of pseudo labels $\{(y^\mathrm{NMS},B^\mathrm{NMS})\}$.
% we obtain the features $e^I$ and $e^{RI}$ generated by detector and generator respectively, the set of abnormalities $\{({y^\mathrm{NMS}},{B^\mathrm{NMS}})\}$ detected in $x^u$ after SA-NMS, and the predictions \textcolor{red}{$\{\hat{y}^R\}$} classified by $F^R$.
%We first concatenate the features $e=[e_I, e_R]$ as the enhanced input of the detector.
Finally, we only keep the pseudo labels whose categories are in the generator-predicted abnormalities:
\begin{equation}\label{Eqn. filter-det}
  \{(y^p,B^p)\} = \left. \left\{\left({y^\mathrm{NMS}_i},{B^\mathrm{NMS}_i}\right) \right| {y^\mathrm{NMS}_i} \in \{\hat{y}^R\}\right\}.
\end{equation}
Eventually, the student detection model is trained with $\{(y^p,B^p)\}$ as the pseudo labels of weakly labeled images (Eqn. (\ref{eq:pld})).
%Finally, we utilize Eqn.  (\ref{eq:pld}) to train the student detection model $F_s^I$ using the set of annotated data $\{(y^p,B^p)\}$ and the enhanced feature $e$.

\subsubsection{Co-evolutionary Strategy}
Ideally, one should use an optimal report generation model to refine and supplement the abnormality detection task and vice versa.
However, the two models mutually depend on each other in a circle.
To solve this dilemma, we implement an alternative co-evolution strategy to improve the abnormality detection and report generation cyclically in iterations.
As shown in Fig. \ref{fig:co-evolution}, the $k$\textsuperscript{th} iteration student detection model $F_{s,k}^I$ is distilled from the teacher $F_{t,k-1}^I$, whose pseudo labels are refined and input features are enhanced by the prediction of the frozen generation model $F_{k}^R$ via GIP. 
{\color{blue}Subsequently, $F_{s,k}^I$ is frozen and used to} 1) help train the $(k+1)$\textsuperscript{th} generation model $F_{k+1}^R$ via DIP, and 2) {\color{blue}serve as the teacher detection model in the next iteration: $F_{s,k}^I\rightarrow F_{t, k}^I$.\footnote{The initial teacher $F_{t,0}^I$ is obtained by training on the labeled data only.}
This way, the learned knowledge is passed down from one iteration to the next.
Meanwhile, the generation and student detection models are reborn with random initialization in each iteration.
This born-again training strategy has proved effective in the teacher-student paradigm \cite{furlanello2018born}, by learning from a more knowledgeable teacher in each iteration while avoiding getting stuck by suboptimal solutions.}
Thus, the co-evolution continues to optimize the detector and generator cyclically with cross-task mutual promotion.
After training, we only need the final student detection model $F_{s,K}^I$ and report generation model $F_{K+1}^R$ for upcoming test CXRs.

\subsection{Handling Normal Cases}
{\color{blue}
In clinical settings, most of the daily cases examined will be normal.
Thus, any framework to properly integrate into clinical routine should handle normal cases.
Let us assume that the student detection model does not detect any abnormality in a weakly labeled CXR image $x$. 
In that case, we use the entire input image as the detection output $(\hat{y}, \hat{B})$, where the category $\hat{y}=0$ indicates the background and the bounding box $\hat{B}=(0, 0, W, H)$ indicates the whole image. 
Then, $(\hat{y},\hat{B})$ is used for report generation. 
Specifically, a token $e^R$ is extracted from $x$ and a location embedding $e^{loc}$ from $(0, 0, W, H)$. 
Next, $e^R$ and $e^{loc}$ are fed into the report generation model. 
In addition, if neither the teacher nor the student detection model detects any abnormality in $x$, $x$ will be excluded from the loss computation in the semi-supervised learning (Eqn. (\ref{eq:pld})). 
On the other hand, if the report generator does not predict any abnormality, $x$ will also be excluded from the semi-supervised learning regardless of the detection models’ output. 
This is because only the common abnormalities identified by both the detection and generation models would be kept (Eqn. (\ref{Eqn. filter-det})).
It is worth noting that $x$ is still used to train the generation model. Thus, our co-evolving system can handle normal cases even if no disease is detected / predicted by either model. 

Meanwhile, it is also possible that abnormalities are detected for a normal CXR by the detection models, \textit{i.e.}, false detections.
In that case, it is likely that the report generator does not predict any abnormality in the regions identified by the detection models. 
Thus, the false detections would be removed by Eqn. (\ref{Eqn. filter-det}), as described above. 
Therefore, this strategy effectively reduces false-positive pseudo labels for the semi-supervised training, which in turn makes the student detection model produce fewer false positives.
}

\subsection{Inference}
{\color{blue}Given a test CXR $x$, we use the student detection model $F_{s,K}^I$ and report generation model $F_{K+1}^R$ for inference as described below.
First, $x$ is fed to $F_{s,K}^I$, yielding the abnormality detection result: $\{(\hat{y},\hat{B})\}=F_{s,K}^I(x)$.
Then, the abnormality tokens $\{e^R\}$ and location embeddings $\{e^{loc}\}$ are extracted from $\{(\hat{y},\hat{B})\}$.
Finally, $\{e^R\}$ and $\{e^{loc}\}$ are fed to $F_{K+1}^R$ to generate the report for $x$.}

% Jinghan original:
% \todo{At inference, given the student detection model $F_{s,K}^I$ and report generation model $F_{K+1}^R$, for a test CXR $x$, we first generate abnormality predictions by: $\{(\hat{y},\hat{B})\}=F_{s,K}^I(x)$. Then, we obtain the corresponding abnormality token $e^R$ and location embedding $e^{loc}$. Thus, we generate the report by: $\{\hat{y}^R\}, \{\hat{r}_{1}, \ldots, \hat{r}_{T}\}=F_{K+1}^R\left({e^{cls}}, \{[{e^{R}}, {e^{loc}}]\}\right).$}

\section{Experiments}
\subsection{Datasets and Evaluation Metrics}
We conduct experiments on the chest radiography dataset MIMIC-CXR \cite{johnsonmimic,johnson2019mimic} with the detection annotations provided by MS-CXR \cite{boecking2022making} and {\color{blue}a pneumo-disease CXR (PD-CXR) dataset focused on pneumonia and pneumothorax \cite{tam2020weakly}}.
MIMIC-CXR is a large publicly available dataset of CXRs and free-text radiology reports. 
MS-CXR provides bounding box annotations for part of the CXRs in MIMIC-CXR (1,026 samples).
Specifically, MIMIC-CXR includes 14 categories of anatomical abnormalities for multi-label classification of the reports, while MS-CXR includes bounding box annotations for eight different cardiopulmonary radiological findings.
For consistency, we exclude samples in MIMIC-CXR that have abnormality labels outside the eight categories of MS-CXR, leaving 112,425 samples.
Thus, in our experiments, 1,026 samples are fully labeled with bounding boxes, and the rest are not.
% We will refer to the combined dataset of fully labeled and weakly labeled samples as MS-CXR-E (E for extended).
{\color{blue}Similarly, the PD-CXR dataset contains 455 samples with bounding box annotations exclusively for pneumonia and pneumothorax.
To create a separate dataset focused on these two categories, we exclude MIMIC-CXR samples with abnormality labels other than pneumonia and pneumothorax, resulting in 24,753 samples.
For both MS-CXR and PD-CXR,} we respectively split the weakly labeled and fully labeled samples for training, validation, and testing according to the ratio of 7:1:2. 
For abnormality detection, we take the test sets of fully labeled samples for testing.
For report generation, the combination of the test sets of both fully and weakly labeled samples are used for testing. 
We focus on frontal views in this work.

The mean average precision (mAP) \cite{everingham2009pascal} with various intersection over union (IoU) thresholds is employed to evaluate the performance of abnormality detection in CXR.
Specifically, the thresholds 0.25, 0.5, and 0.75 are used for MS-CXR following \cite{xu2021end}, whereas {\color{blue}0.1, 0.3, and 0.5 are used for PD-CXR following \cite{tam2020weakly}}.
Six commonly used language-efficacy metrics are employed to evaluate the quality of generated reports: BLEU (1- to 4-gram) \cite{papineni2002bleu},  METEOR \cite{banerjee2005meteor}, and ROUGE \cite{lin2004rouge}.
{\color{blue}The Wilcoxon signed-rank test is used for statistical significance testing in report generation, where a $p$-value below 0.05 is considered significant.}
In addition, we assess the clinical efficacy of generated reports by calculating the area under the curve (AUC) using report-extracted abnormalities by CheXBert \cite{smit2020chexbert}.

\subsection{Implementation}
The PyTorch \cite{paszke2019pytorch} framework (1.4.0) is used for experiments.
Our results are obtained on a single machine with four NVIDIA V100 GPUs.
For abnormality detection, we employ RetinaNet \cite{lin2017focal} with ResNet-101 \cite{he2016deep} pretrained on ImageNet \cite{krizhevsky2012imagenet} and FPN \cite{lin2017feature} as the backbone. 
We resize all images to 512$\times$512 pixels and use a batch size of 16. 
Data augmentation, including random cropping and flipping, is performed.
The ratio of labeled to \textcolor{blue}{weakly labeled} samples in each mini-batch during the semi-supervised training is empirically set to 2:1.
Other implementation details and hyper-parameters (\textit{e.g.}, optimizer and learning rate) follow the official settings in \cite{lin2017focal}.
For report generation, we follow the baseline model R2Gen\cite{chen2020generating} unless otherwise specified.
%R2GenCMN\cite{chen2022cross}
A ResNet-101 \cite{he2016deep} pretrained on ImageNet \cite{krizhevsky2012imagenet} is adopted as the projection network $f^R_{prj}$.
We optimize the model design (\textit{e.g.}, hyperparameters, and ablation studies) on the validation datasets and keep the test datasets for final performance evaluation only.
% We evaluate the models on the validation dataset for model optimization.
% \textbf{
% You use a ResNet for image patch projection. Do the baseline models use it, too? (Yes)
Unless otherwise stated, {we evolve the detection and generation models for three iterations} and train both models for 20 epochs in each iteration (including initial training of the teacher model).

\begin{table}[t]
\centering
\caption{The effect of the pseudo-label threshold on the baseline TSD model \cite{hinton2015distilling} (Eqn. (\ref{eq:pld})) on the validation set of MS-CXR, using mAP (\%) with the IoU thresholds of 0.25, 0.5, and 0.75. 
The best results are shown in bold.}
\begin{tabular}{c|ccc}
\hline
Threshold & @0.25 & @0.5 & @0.75 \\ \hline
0.70      & 36.65                           & 31.07                           & 15.50                           \\
0.80      & \textbf{37.41}                           & 32.70                           & 16.97                           \\
0.90      & {37.31} & \textbf{32.76} & \textbf{17.96} \\
0.95      & 37.05                           & 32.11                           & 16.42                           \\ \hline
\end{tabular}
\label{tab:ablation_threshold}
\end{table}

\subsection{Hyperparameter Validation}
\subsubsection{Threshold for Pseudo-label Filtering}
In TSD, the threshold for pseudo-label filtering is an important hyperparameter.
Only candidates with prediction possibilities higher than a threshold are retained.
Therefore, we empirically study the effect of varying the threshold on the validation set of MS-CXR.
As shown in Table \ref{tab:ablation_threshold}, setting the threshold = 0.9 achieves the highest mAPs @0.5 and @0.75, and the second highest mAP@0.25.
Hence, we fix the pseudo-label filtering threshold to 0.9 for subsequent evaluation and comparison with other methods on the test sets.

% \begin{figure}[htbp]
% \centering
% \begin{minipage}[t]{0.48\textwidth}
% \centering
% \includegraphics[width=6cm]{{LaTeX/Detection.png}
% \caption{World Map}
% \end{minipage}
% \begin{minipage}[t]{0.48\textwidth}
% \centering
% \includegraphics[width=6cm]{LaTeX/Generation.png}
% \caption{Concrete and Constructions}
% \end{minipage}
% \end{figure}
\begin{figure}
  \centering
  \includegraphics[scale=0.375]{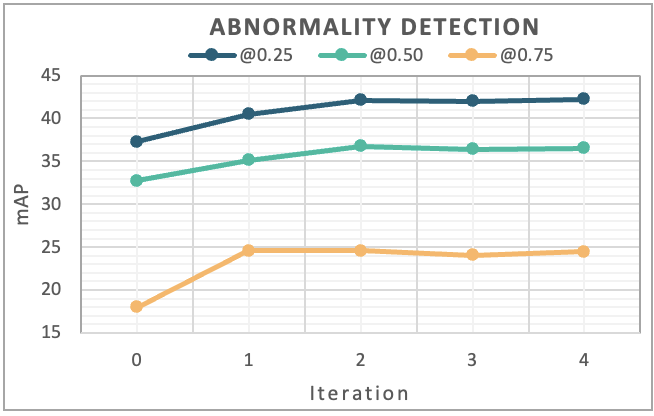}
  % \hspace{0.0in}
  \includegraphics[scale=0.375]{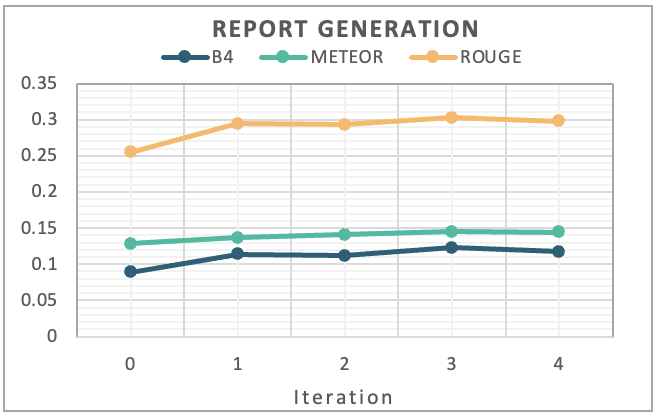}
  \caption{Performance of the detection and report generation models as a function of iterations on the validation set of MS-CXR.}
\label{fig:generation}
\end{figure}

\begin{table*}[]
\centering
\caption{Abnormality detection results on the test sets, using mAP (\%) with various IoU thresholds. 
The best results, excluding semi-oracle, are shown in bold.}
% \begin{adjustbox}{width=0.9\linewidth}
\begin{tabular}{c|l|ccc|ccc}
\hline
 \multirow{2}{*}{Supervision}&\multirow{2}{*}{Method} & \multicolumn{3}{c|}{MS-CXR} & \multicolumn{3}{c}{{\color{blue}PD-CXR}}\\ \cline{3-8}
 &       & @0.25                      & @0.50                      & @0.75  & \blue{@0.10} & \blue{@0.30} & \blue{@0.50}  \\ \hline
\multirow{2}{*}{Weak} &CAM \cite{zhou2016learning}         & 20.47                      & 11.20                     & 3.05  &\blue{27.41} & \blue{25.87} & \blue{10.02} \\
&AGXNet \cite{yu2022anatomy}      & 29.96                      & 15.62                     & 7.44   & \blue{28.10} & \blue{25.92} & \blue{12.78}\\ \hdashline
Full & RetinaNet \cite{lin2017focal}       & 37.91                      & 32.84                      & 19.21  & \blue{31.72}& \blue{28.11}& \blue{16.55}\\ \hdashline
Unified & GPV \cite{gupta2022towards}         & 37.89                      & 32.41                      & 20.55  & \blue{33.78} & \blue{31.95} & \blue{19.09}\\ 
vision-language &GLIPv2 \cite{zhang2022glipv2}      & 36.38                      & 30.91                      & 16.32  &\blue{32.35} &\blue{30.33}& \blue{17.90}\\ \hdashline
&TSD \cite{hinton2015distilling}         & 38.29                      & 33.95                      & 19.51  & \blue{32.93} & \blue{30.07} & \blue{17.15}\\
\multirow{2}{*}{Semi}&STAC \cite{sohn2020simple}        & 39.26                      & 35.01                      & 23.90  & \blue{31.18} & \blue{31.05} & \blue{18.94}\\
& LabelMatch \cite{chen2022label}  & 39.92                      & 36.40                      & 24.06  & \blue{33.85} & \blue{31.90} & \blue{19.17}\\
&Soft Teacher \cite{xu2021end} & 40.17                      & 36.59                      & 24.78  &\blue{34.05}&\blue{32.18} & \blue{19.32}\\ \hdashline
\multirow{2}{*}{Semi + weak} & {\color{blue}CEIRD \cite{sun2023you}}         & \blue{41.93}                      & \blue{37.20}                      & \blue{25.12}  &\blue{34.19}&\blue{32.77}&\blue{20.01}\\ 
 & CoE-DG (Ours)         &   \textbf{42.24}                      & \textbf{37.89}                      & \textbf{25.93}& \textbf{\color{blue}34.71} & \blue{\textbf{33.14}} & \blue{\textbf{20.19}}\\ \hline
Semi + weak & \textit{Semi-oracle}  & 43.70                               & 38.05                               & 26.15  & \blue{35.19} & \blue{33.49} & \blue{20.55}\\ \hline
\end{tabular}
% \end{adjustbox}
\label{tab:det_res}
\end{table*}

\begin{table*}[]\color{blue}
\caption{Abnormality detection results of our framework for each category in the MS-CXR dataset on the test set, using mAP (\%) with the IoU threshold  0.5.
The best results, excluding semi-oracle, are shown in bold.}
\label{tab:det-clswse}
\setlength{\tabcolsep}{0.8mm}
\begin{adjustbox}{width=\linewidth}
\begin{tabular}{c|l|cccccccc|c}
\hline
Supervision                                                                & Method                                           & Cardiomegaly & Lung opacity & Edema & Consolidation & Pneumonia & Atelectasis & Pneumothorax & Pleural effusion & Average                             \\ \hline
\multirow{2}{*}{Weak}                                                      & CAM \cite{zhou2016learning}      & 58.56                            & 5.30                             & 0.00                      & 8.20          & 10.09     & 3.27        & 1.38         & 4.82             & 11.20                           \\
                                                                           & AGXNet \cite{yu2022anatomy}      & 67.44                            & 1.46                             & 7.87                      & 15.83         & 9.62      & 1.21        & 6.04         & 15.53            & 15.62                           \\
\hdashline 
Full                                              & RetinaNet \cite{lin2017focal}                  & 79.03                            & 18.49                            & 66.68                     & 44.62         & 12.88     & 27.33       & 8.61         & 3.71             & 32.84                           \\
\hdashline Unified                                           & GPV \cite{gupta2022towards}                    & \textbf{82.64}                            & 8.04                             & 67.32                     & 42.14         & 13.56     & 22.75       & 10.35        & 12.52            & 32.41                           \\
vision-language                                                            & GLIPv2 \cite{zhang2022glipv2}    & 77.18                            & 15.04                            & 61.01                     & 37.34         & \textbf{15.03}     & 17.20       & 10.95        & 13.58            & 30.91                           \\
\hdashline                                                  & TSD \cite{hinton2015distilling}                 & 79.03                            & 18.49                            & 66.68                     & 44.62         & 12.88     & 27.33       & 8.91         & 13.71            & 33.95                           \\
\multirow{2}{*}{Semi}                                                      & STAC \cite{sohn2020simple}       & 81.52                            & 21.63                            & 66.82                     & 42.37         & 14.30     & 30.54       & 10.35        & 12.99            & 35.01                           \\
                                                                           & LabelMatch \cite{chen2022label}  & 81.95                           & 23.54                            & 66.41                     & 43.15         & 13.58     & 34.12       & 14.86        & 13.47            & 36.40                           \\
                                                                           & Soft Teacher \cite{xu2021end}    & 80.22                            & 23.18                            & 65.79                     & 43.43         & 13.71     & 33.66       & 16.96        & 15.80            & 36.59                           \\
\hdashline\multirow{2}{*}{Semi + weak} & CEIRD \cite{sun2023you}                                              & 81.94 & 23.15 & 66.90 & 44.10 & 14.21 & \textbf{35.89} & 16.34 & 15.10 & 37.20  \\
                                                                           & CoE-DG (Ours)                    & 81.92 & \textbf{24.76} & \textbf{67.65} & \textbf{44.72} & 14.84 & 35.48 & \textbf{16.98} & \textbf{16.73} & \textbf{37.89}  \\ \hline
Semi + weak                                                                & \textit{Semi-oracle}             & 82.52 & 24.18 & 66.16 & 42.78 & 19.01 & 34.02 & 18.08 & 17.62 & 38.05                           \\ \hline
\end{tabular}
\end{adjustbox}
\end{table*}

% \begin{figure}[]
% \centering
% \includegraphics[width=0.5\linewidth]{LaTeX/Detection.png}
% \includegraphics[width=0.5\linewidth]{LaTeX/Generation.png}
% \caption{Performance of the detection and report generation models as a function of iterations.}
% \label{fig:co-evolution}
% \end{figure}

\subsubsection{Number of Co-evolution Iterations}
Furthermore, we conduct experiments on the validation set of MS-CXR to empirically determine the optimal number of iterations for co-evolution. 
The results are shown in Fig. \ref{fig:generation}.
{For abnormality detection, the 0\textsuperscript{th} iteration is a fully supervised model trained on labeled data only (\textit{i.e.}, the initial model $F^I_{t,0}$).
For report generation, the 0\textsuperscript{th} iteration is a baseline model (\textit{i.e.}, R2Gen) that is included for reference only but not needed in our framework.
As we can see, the detection and generation models improve in the first three and four iterations, respectively,} and then remain stable, confirming that both models promote each other with the co-evolution strategy. 
Therefore, we select the 2\textsuperscript{nd}-iteration detection model and 3\textsuperscript{rd}-iteration generation model for comparison with other methods.

\begin{table*}[t]
\centering
\caption{Report generation results on the test sets. BLEU-N denotes N-gram score of BLEU \cite{papineni2002bleu}. 
The best results are shown in bold. 
{\color{blue}*, $^\dagger$ and $^\ddag$ indicate statistically significant differences from R2GenCMN+Ours, R2Gen+Ours, and ORGan+Ours, respectively, for pairwise comparison with the Wilcoxon signed-rank test ($p<0.05$).}
% {\color{red}Better rerun the ORGan experiments more times to obtain more convincing results.}
}
\setlength{\tabcolsep}{.75mm}
\begin{adjustbox}{width=\textwidth}
\begin{tabular}{l|cccccc|c|cccccc|c}
\hline
\multirow{2}{*}{Method} & \multicolumn{7}{c|}{MS-CXR+MIMIC-CXR} & \multicolumn{7}{c}{{\color{blue}PD-CXR+MIMIC-CXR}}\\ \cline{2-15}
               & {BLEU-1}     & {BLEU-2}     & {BLEU-3} & {BLEU-4} & {METEOR} &{ROUGE} & {AUC}  & {\color{blue}BLEU-1}     & {\color{blue}BLEU-2}     & {\color{blue}BLEU-3} & {\color{blue}BLEU-4} & {\color{blue}METEOR} &{\color{blue}ROUGE} & {\color{blue}AUC}                 \\ \hline
{\color{blue}LLaVA-Med \cite{li2024llava}}         & {\color{blue}0.1030*}   &{\color{blue}0.0601*}   &{\color{blue}0.0392*}    & {\color{blue}0.0265*}                      & {\color{blue}0.0743*}                           &  {\color{blue}0.1324*}                          &  {\color{blue}0.51}                &{\color{blue}0.1387*} & {\color{blue}0.0894*} & {\color{blue}0.0623*} & {\color{blue}0.0453*} & {\color{blue}0.0824*} & {\color{blue}0.1712*}& {\color{blue}0.58} \\
{\color{blue}XrayGPT\cite{thawkar2023xraygpt}} &  {\color{blue}0.2346*} & {\color{blue}0.1068*} & {\color{blue}0.0528*} & {\color{blue}0.0277*} & {\color{blue}0.0931*} & {\color{blue}0.1938*}                          &  {\color{blue}0.62}          & {\color{blue}0.2105*} & {\color{blue}0.1002*} & {\color{blue}0.0436*} & {\color{blue}0.0163*} & {\color{blue}0.0742*} & {\color{blue}0.1612*} & {\color{blue}0.64}       \\
{\color{blue}CheXagent \cite{chen2024chexagent}}& {\color{blue}0.1730*}   &{\color{blue}0.1301*}   &{\color{blue}0.1092*}    & {\color{blue}0.0965*}                      & {\color{blue}0.0813*}                           &  {\color{blue}0.1945*}                          &  {\color{blue}0.61}          & {\color{blue}0.1972*} & {\color{blue}0.0967*} & {\color{blue}0.0665*} & {\color{blue}0.0484*} & {\color{blue}0.0907*} & {\color{blue}0.1504*} & {\color{blue}0.65}       \\
\hdashline
GPV \cite{gupta2022towards}           & {0.2865*}                           & {0.1649*}                          & {0.1019*}                      &          {0.0679*}             &    {0.1044*}                  &  {0.2419*}                    &  {0.60}                   & {\color{blue}0.2165*} & {\color{blue}0.1289*} & {\color{blue}0.0848*} & {\color{blue}0.0608*} & {\color{blue}0.0878*} & {\color{blue}0.1905*} & {\color{blue}0.67} \\
GLIPv2 \cite{zhang2022glipv2}         & {0.3057*}  & {0.1762*}  & {0.1064*}   & {0.0698*}                       &   {0.1064*}                         &    {0.2390*}                        &   {0.60}          & {\color{blue}0.2141*} & {\color{blue}0.1259*} & {\color{blue}0.0808*} & {\color{blue}0.0575*} & {\color{blue}0.0833*} & {\color{blue}0.1896*} & {\color{blue}0.67}         \\ 
\hdashline
R2Gen \cite{chen2020generating}         & {0.3183}$^\dagger$ & {0.1866}$^\dagger$ & {0.1210$^\dagger$}                & {0.0834$^\dagger$}                 & {0.1263$^\dagger$}                     & {0.2566$^\dagger$}                     & {0.63} & {\color{blue}0.2220$^\dagger$} & {\color{blue}0.1322$^\dagger$} & {\color{blue}0.0858$^\dagger$} & {\color{blue}0.0611$^\dagger$} & {\color{blue}0.0919} & {\color{blue}0.2011} & {\color{blue}0.65} \\
+Ours    & \textbf{0.3423}                     & \textbf{0.2123}                     & \textbf{0.1425}                 & \textbf{0.1019}                 & \textbf{0.1409}                     & \textbf{0.2733}                     & \textbf{0.69}                    &{\color{blue}\textbf{0.2472}} & {\color{blue}\textbf{0.1466}} & {\color{blue}\textbf{0.0954}} & {\color{blue}\textbf{0.0678}} & {\color{blue}\textbf{0.0951}} & {\color{blue}\textbf{0.2034}} & {\color{blue}\textbf{0.70}} \\ \hdashline%0.7689
R2GenCMN  \cite{chen2022cross}     & {0.3187*}                     & {0.1967*}                     & {0.1317*}                 & {0.0942*}                 & {0.1306*}                     & {0.2690*}                     & {0.64} &{\color{blue}0.2458*} & {\color{blue}0.1445*} & {\color{blue}0.0933*} & {\color{blue}0.0660*} & {\color{blue}0.1010} & {\color{blue}0.2363*}  & {\color{blue}0.66}\\
+Ours & \textbf{0.3434}                     & \textbf{0.2210}                     & \textbf{0.1616}                 & \textbf{0.1184}                 & \textbf{0.1393}                     & \textbf{0.2854}                     & \textbf{0.69}                &{\color{blue}\textbf{0.2608}} & {\color{blue}\textbf{0.1578}} & {\color{blue}\textbf{0.1056}} & {\color{blue}\textbf{0.0776}} & {\color{blue}\textbf{0.1119}} & {\color{blue}\textbf{0.2468}} & {\color{blue}\textbf{0.68}}\\ 
\hdashline%0.7689
{\color{blue}ORGan  \cite{hou2023organ}}     & {\color{blue}0.3724$^\ddag$}                     & {\color{blue}0.2218$^\ddag$}                     & {\color{blue}0.1557}                 & {\color{blue}0.1185$^\ddag$}                 & {\color{blue}\textbf{0.1525}}                     & {\color{blue}0.2913}                     & {\color{blue}0.68} & {\color{blue}0.2924$^\ddag$} & {\color{blue}0.1782$^\ddag$} & {\color{blue}0.1128$^\ddag$} & {\color{blue}\textbf{0.1044}} & {\color{blue}0.1224$^\ddag$} & {\color{blue}{0.2437}}& {\color{blue}0.68} \\
{\color{blue}+Ours} & {\color{blue}\textbf{0.3810}}                     & {\color{blue}\textbf{0.2275}}                     & {\color{blue}\textbf{0.1562}}                 & {\color{blue}\textbf{0.1191}}                 & {\color{blue}0.1519}                     & {\color{blue}\textbf{0.2933}}                     & {\color{blue}\textbf{0.70}}              & {\color{blue}\textbf{0.3094}} & {\color{blue}\textbf{0.1841}} & {\color{blue}\textbf{0.1210}} & {\color{blue}0.0974} & {\color{blue}\textbf{0.1296}} & \textbf{\color{blue}0.2449}& {\color{blue}\textbf{0.70}}  \\
\hline%0.7639
\end{tabular}
\label{tab:rep_res}
\end{adjustbox}
\end{table*}

\begin{figure}[t]
\centering
\includegraphics[width=0.9\linewidth]{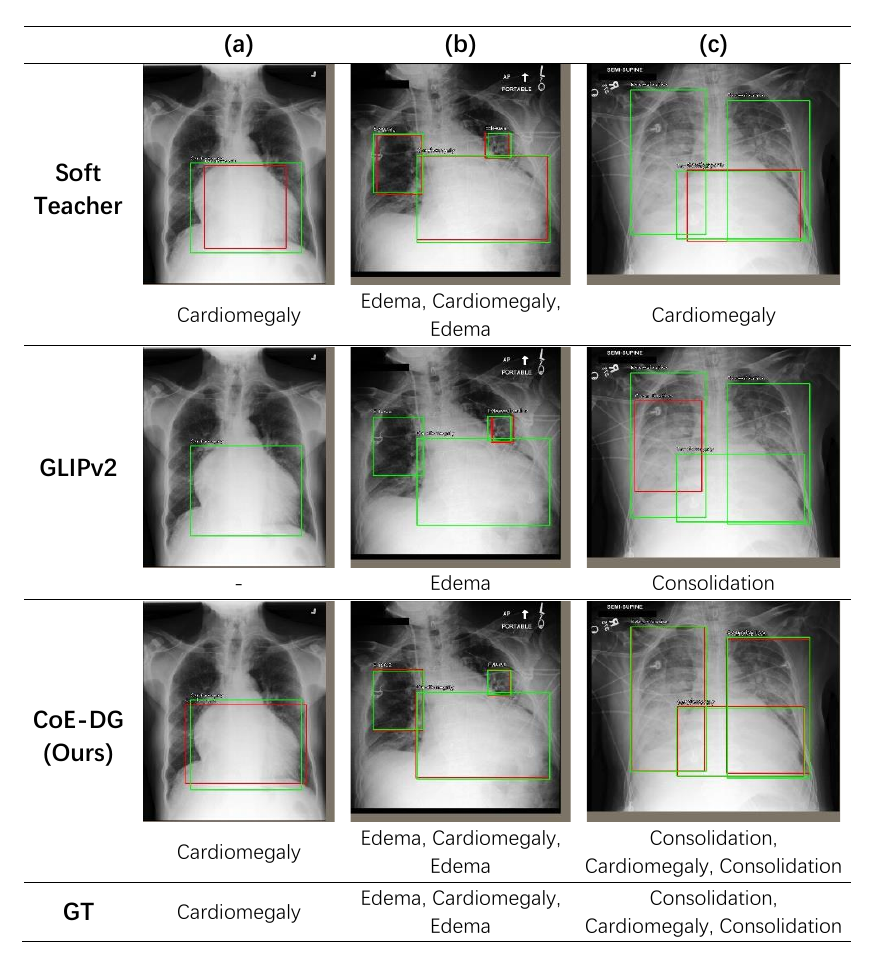}
\caption{Visualization of the detection results by Soft Teacher \cite{xu2021end}, GLIPv2 \cite{zhang2022glipv2} and ours.
The green and red boxes are the ground truth (GT) and predictions, respectively.
The texts below images indicate abnormality categories of the boxes from left to right, and ``-'' indicates a failure to detect an abnormality.
}
\label{fig:vis-det}
\end{figure}

\subsection{Comparison with State-of-the-Art (SOTA) Object Detection Methods}
For abnormality detection in CXR, we compare our proposed CoE-DG framework with several up-to-date detection methods, including:
\begin{itemize}
    \item Weakly supervised by the abnormality categories $\{y^l\}$ and reports in $D^l$: CAM \cite{zhou2016learning} (locating objects based on class activation maps) and AGXNet \cite{yu2022anatomy} (aiding CAM-based localization with report representations);
    \item A vanilla object detection model fully supervised by $\{(y^l, B^l)\} \in D^l$ but not the reports: RetinaNet \cite{lin2017focal};
    \item Unified vision-language models first pretrained by both $\{(y^l, B^l)\}$ and reports in $D^l$ and then fine-tuned fully supervised by $\{(y^l, B^l)\}$: GPV \cite{gupta2022towards} and GLIPv2 \cite{zhang2022glipv2};
    \item Semi-supervised on $D^l \cup D^w$ but without utilizing the reports: baseline teacher-student pseudo-label distillation  \cite{hinton2015distilling} (TSD; see Eqn.  (\ref{eq:pld})), and three SOTA ones: STAC \cite{sohn2020simple}, LabelMatch \cite{chen2022label}, and Soft Teacher \cite{xu2021end}.
\end{itemize}
In comparison, our framework fully uses all the data and both strong and weak supervision signals offered by the training sets $D^l \cup D^w$.
{\color{blue}We also include a preliminary, conference version, of our work~\cite{sun2023you} in the comparison.}

{\color{blue}For implementation, we reimplement CAM following the original recipe as described in \cite{zhou2016learning}.
For other compared methods, we use the official codes and optimize the performance using the validation data splits following the optimal practices suggested by the original authors.
Notably, our framework uses the RetinaNet baseline method as its detection model.}

%weakly supervised: CAM \cite{zhou2016learning} (locating objects based on class activation maps) and AGXNet \cite{yu2022anatomy} (aiding CAM-based localization with report representations), fully supervised on labeled training data only (Sup.), baseline semi-supervised via teacher-student pseudo-label distillation  (TSD; see Eqn.  (\ref{eq:pld})) \cite{hinton2015distilling}, three SOTA semi-supervised: STAC \cite{sohn2020simple}, LabelMatch \cite{chen2022label}, and Soft Teacher \cite{xu2021end}, and two unified vision-language models: GPV \cite{gupta2022towards} and GLIPv2 \cite{zhang2022glipv2}. 

The results are shown in Table~\ref{tab:det_res}, from which we make the following observations. 
First, all fully supervised (RetinaNet and unified models) and semi-supervised methods outperform the weakly supervised by large margins.
% proving the efficacy of using annotations. 
Second, the unified vision-language models are comparable to the fully supervised RetinaNet on MX-CXR {\color{blue}and marginally better on PD-CXR}.
%the unified methods are generally worse than both fully and semi-supervised methods.
Third, all semi-supervised methods outperform the fully supervised. 
% by more than 0.30\%
Finally, our CoE-DG achieves the best performance for the mAPs evaluated at three different IoU thresholds, outperforming the second-best method Soft Teacher (excluding our preliminary version \cite{sun2023you}) by up to 2.07\%.
In addition, we evaluate a semi-oracle of our method, where the ground truth multi-labels provided in MIMIC-CXR are used for GIP instead of the generation model's prediction.
As we can see, our method is marginally short of the semi-oracle, \textit{e.g.}, 37.89\% versus 38.05\% for mAP@0.5 on MS-CXR.
{\color{blue}For reference, Table \ref{tab:det-clswse} presents the mAP@0.5 of our method for each abnormality category in the MS-CXR dataset.}

Fig. \ref{fig:vis-det} visualizes the detection results of Soft Teacher \cite{xu2021end}, GLIPv2 \cite{zhang2022glipv2}, and our method.
As observed, Soft Teacher and GLIPv2 sometimes miss the abnormalities, whereas ours detects all abnormalities and locates them more accurately.

\begin{table}[t]
\centering
\caption{Abnormality detection results on the test set of MS-CXR with different amounts of fully labeled data (25\% and 50\% of $D^l$), 
%Comparison with SOTA on the test set with different annotation data ratios (25\% and 50\%), 
using mAP (\%) with the IoU thresholds of 0.25, 0.5, and 0.75. 
%Note that percentages refer to the proportion of annotated data. 
The best results are shown in bold.}
\setlength{\tabcolsep}{1.6mm}
% \begin{adjustbox}{width=0.9\linewidth}
\begin{tabular}{l|ccc|ccc}
\hline
             & \multicolumn{3}{c|}{25\%} & \multicolumn{3}{c}{50\%}  \\ \hline
Method                           & @0.25   & @0.5   & @0.75  & @0.25   & @0.5   & @0.75   \\ \hline
RetinaNet \cite{lin2017focal}                              & 21.18        &  17.07      & 8.95       & 26.63        &  21.58      & 7.46          \\ \hdashline
TSD \cite{hinton2015distilling}  & 21.50        &  17.16      & 9.06       & 28.16        &  23.91      & 11.89         \\
STAC \cite{sohn2020simple}       & 22.75        &  18.27      & 9.14       & 29.13        &  24.56      & 13.92         \\
LabelMatch \cite{chen2022label}  & 25.53        &  20.58      & 6.46       & 30.85        &  25.74      & 14.66         \\
Soft Teacher \cite{xu2021end}    & 27.07        &  21.16      & 10.42      & 31.52        &  26.25      & 14.79         \\ \hdashline
CoE-DG (Ours)                             & \textbf{27.64} &  \textbf{22.81}   & \textbf{11.37} & \textbf{33.05} &  \textbf{27.39}   & \textbf{15.21}     \\ \hline
\end{tabular}
% \end{adjustbox}
\label{tab:det_ratio}
\end{table}

Furthermore, we conduct a control experiment on the amount of fully labeled data available for training to investigate the label efficiency of the compared methods. 
%a comparative experiment on the test set using different fully annotated data ratios to evaluate the performance of our method with fewer box annotations.
The results are shown in Table \ref{tab:det_ratio}. 
Our CoE-DG consistently outperforms others by notable margins (0.57\%-1.65\%).

% \begin{table*}[t]
% \caption{Abnormality detection results on the test data, using mAP (\%) with the IoU thresholds of 0.25, 0.5, and 0.75.
% TSD: teacher-student distillation.}
% \begin{adjustbox}{width=\textwidth}
% \begin{tabular}{l|cc;{2pt/3pt}c;{2pt/3pt}cccc;{2pt/3pt}ccc;{2pt/3pt}c}
% \hline
% {mAP}           & {CAM \cite{zhou2016learning}} & {AGXNet \cite{yu2022anatomy}} & {Sup.} & {TSD \cite{hinton2015distilling}} & {STAC \cite{sohn2020simple}} & {LabelMatch \cite{chen2022label}} & {Soft Teacher \cite{xu2021end}} & GPV\cite{gupta2022towards} & GLIPv2\cite{zhang2022glipv2} & {Ours}     & {Semi-oracle} \\ \hline
% {@0.25} & {20.47}  & {29.96}  & {37.91} & {38.29} & {39.26} & {39.92} & {40.17} & 37.89 & 36.38 & {\textbf{42.24}} & 43.70  \\ 
% {@0.5}  & {11.20}  & {15.62}  & {32.84} & {33.95} & {35.01} & {36.40} & {36.59} & 32.41 & 30.91 & {\textbf{37.89}} & 38.05  \\ 
% {@0.75} & {3.05}   & {7.44}   & {19.21} & {19.51} & {23.90} & {24.06} & {24.78} & 20.55 & 16.32 & {\textbf{25.93}} & 26.15  \\ \hline
% \end{tabular}
% \end{adjustbox}
% \label{tab:det_res}
% \end{table*}

\subsection{Comparison with SOTA Report Generation Methods}\label{sec:compare}
For report generation, we compare our CoE-DG framework with several methods, including task-specific report generation models (R2Gen \cite{chen2020generating}, R2GenCMN \cite{chen2022cross}, and {\color{blue}ORGan \cite{hou2023organ}}) and unified vision-language models (GPV \cite{gupta2022towards} and GLIPv2 \cite{zhang2022glipv2}). 
The task-specific models are trained on $D^w \cup D^l$ but only using the CXR images and accompanying reports.
The unified models are first pretrained on $D^l$ using images, reports, and abnormality annotations and then fine-tuned with all images and reports in $D^w \cup D^l$.
In contrast, our model is directly trained on $D^w \cup D^l$ using all available images, reports, and abnormality annotations.
We also note that our model is built on top of the task-specific models, \textit{i.e.}, we use them as the report generator $F^R$ in our framework.
{\color{blue}In addition, we also include three medical large multimodal models (MLMMs) for comparison: LLaVA-Med \cite{li2024llava}, CheXagent \cite{chen2024chexagent}, and XrayGPT \cite{thawkar2023xraygpt}. These three MLMMs employed pretrained CLIP \cite{zhang2023biomedclip}, EVA-CLIP \cite{sun2023eva}, and MedCLIP \cite{wang2022medclip}, respectively. }

{\color{blue}For R2Gen, R2GenCMN, ORGan, GPV, and GLIPv2, we use the official codes and optimize the performance using the validation data splits following the optimal practices suggested by the original authors. 
It is worth mentioning that the first three baseline methods, R2Gen, R2GenCMN, and ORGan, also constitute the report generation model in our framework. 
As for the MLMMs, we use the officially released, pretrained models for inference.}

As shown in Table \ref{tab:rep_res}, the task-specific models (R2Gen, R2GenCMN, {\color{blue}and ORGan}) consistently outperform the unified models (GPV and GLIPv2)  by {0.0104--0.0667} in language-efficacy metrics.
%and {\color{red}0.03--0.04} in AUC.
Further still, our CoE-DG framework brings improvements of up to {0.0257} in almost all language-efficacy metrics, demonstrating the validity of our co-evolutionary strategy.
%
% In addition, our method achieves AUCs of 0.76--0.77, demonstrating its ability to classify various abnormalities correctly.
In addition, our method outperforms the task-specific report generation models by 2\%--5\% in AUC, demonstrating its superior clinical efficacy besides linguistic capability.

{\color{blue}As for the MLMMs, we have the following observations. 
First, the language-efficiency metrics for the MLMMs are generally low, among the lowest group of all compared methods. 
Second, while the AUCs of LLaVA-Med are low, those of CheXagent and XrayGPT are much higher---close to the AUCs of R2Gen and R2GenCMN, two task-specific report generation models.
Third, our proposed CoE-DG framework substantially outperforms the MLMMs by 0.0377--0.2780 in language-efficiency metrics and 0.03--0.19 in clinical efficiency metrics.
}

\begin{figure}[t]
\centering
\includegraphics[width=1.0\linewidth]{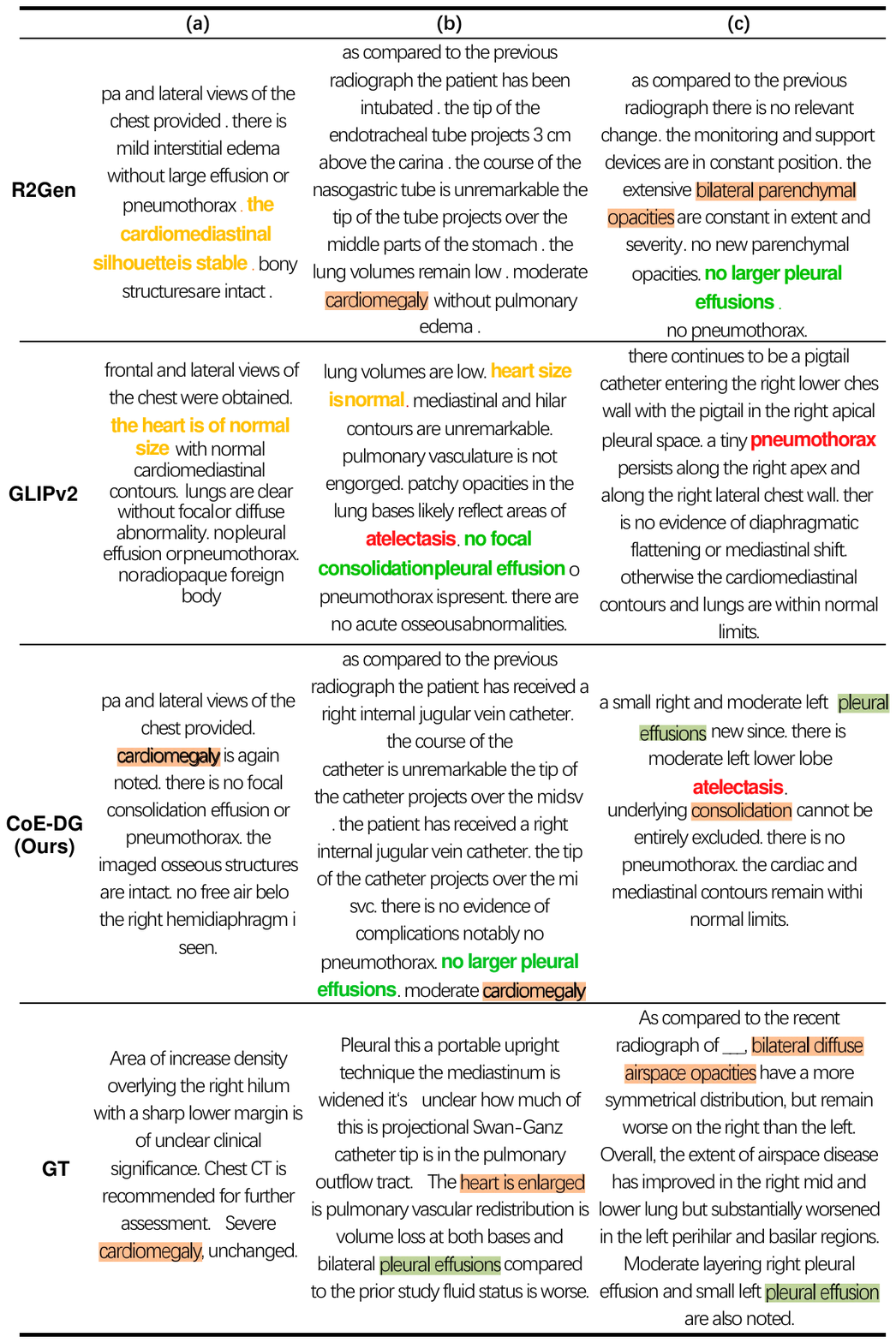}
\caption{Example reports generated by R2Gen\cite{chen2020generating},  GLIPv2\cite{zhang2022glipv2}, our method, and the ground truth (GT).
Texts highlighted with background colors are abnormalities described in GT.
Texts in red indicate false positives, and those in other colors indicate false negatives.
%\textbf{Words highlighted are abnormalities that occur in ground-truth reports. Red words are the false positives, and other bold color words are the false negatives.}
}
\label{fig:vis-gen}
\end{figure}

We visualize example reports generated by R2Gen \cite{chen2020generating}, GLIPv2 \cite{zhang2022glipv2}, CoE-DG (ours), and the ground truth in Fig.~\ref{fig:vis-gen}. 
Our framework describes true positives more often with fewer false positives than other methods.
% \textbf{It is evident that R2Gen sometimes fails to diagnose the disease (a, c). 
% Additionally, most of the sentences generated by GLIPv2 are short and do not provide sufficient information. Furthermore, GLIPv2 usually provides the result of the error diagnosis (b, c). In contrast, our method can diagnose and describe the disease accurately.}

\begin{table}[]\color{blue}
\caption{Abnormality detection results on normal cases from the test set of MS-CXR+MIMIC-CXR.
% {\color{red}The IoU threshold is set to 0.5.}
TNR: true negative rate.}
\label{tab:normal-detect}
\setlength{\tabcolsep}{0.5mm}
\begin{adjustbox}{width=\linewidth}
\begin{tabular}{l|ccccc|c}
\hline
Method   & CAM \cite{zhou2016learning}  & RetinaNet \cite{lin2017focal} & GPV \cite{gupta2022towards}  & LabelMatch \cite{chen2022label} & CoE-DG (ours) & {{Semi-oracle}} \\ \hline
TNR (\%) & 50.00 & 53.33      & 56.67 & 56.67       & 66.67          & 73.33                                                   \\ \hline
\end{tabular}
\end{adjustbox}
\end{table}

\begin{table}[]\color{blue}
\caption{Report generation results on normal cases from the test set of MS-CXR+MIMIC-CXR.
BLEU-N denotes N-gram score of BLEU \cite{papineni2002bleu}. 
TNR: true negative rate.
The best results are shown in bold.
*: statistically significant difference from R2Gen+Ours for pairwise comparison with the Wilcoxon signed-rank test ($p<0.05$).}
\setlength{\tabcolsep}{.7mm}
\begin{adjustbox}{width=\linewidth}
\label{tab:normal-gen}
\begin{tabular}{c|cccccc|c}
\hline
 & BLEU-1 & BLEU-2 & BLEU-3 & BLEU-4 & METEOR & ROUGE  & TNR (\%) \\ \hline
GPV                   & 0.3423* & 0.2129* & 0.1616* & 0.1184* & 0.1428* & 0.2537* & 53.33                   \\
R2Gen                 & 0.3763* & 0.2311* & 0.1503* & 0.1025* & 0.1431* & 0.2725* & 60.00                   \\
R2Gen+Ours            & \textbf{0.4018} & \textbf{0.2435} & \textbf{0.1589} & \textbf{0.1116} & \textbf{0.1559} & \textbf{0.2854} & \textbf{63.33}                   \\ \hline
\end{tabular}
\end{adjustbox}
\end{table}

\begin{table*}[t]
\centering
\caption{Abnormality detection results of ablation study on the validation set of MS-CXR, using mAP (\%) with the IoU thresholds of 0.25, 0.5, and 0.75. "GIP": generator-guided information propagation; "FE": feature enhancement; "PLR": pseudo label refinement; "CoE": co-evolutionary strategy. 
The best results are shown in bold.}
% \begin{adjustbox}{width=0.9\textwidth}
\begin{tabular}{l|ccccc|ccc}
\hline
           & SA-NMS & EMA & GIP-FE & GIP-PLR & CoE & mAP@0.25 & mAP@0.5 & mAP@0.75 \\ \hline
Baseline   &        & &                           &                              &                          & 37.31    & 32.76   & 17.96    \\
Ablation-1 &  \checkmark & & & & & 37.68 & 33.25 & 19.38        \\
\textcolor{blue}{Ablation-2} & & \textcolor{blue}{\checkmark} & & & & \textcolor{blue}{36.68} & \textcolor{blue}{32.05} & \textcolor{blue}{18.42}        \\
Ablation-3 & \checkmark & & \checkmark & & & 38.90 & 34.21 & 20.52  \\
% clstoken  & \checkmark & \checkmark & & & 38.75 & 34.05 & 21.21 \\
Ablation-4 & \checkmark & & \checkmark & \checkmark & & 39.74 & 35.68 & 22.67         \\ 
CoE-DG (ours) & \checkmark & & \checkmark & \checkmark & \checkmark & \textbf{42.13} & \textbf{36.74} & \textbf{24.60}         \\ 
\hline
\end{tabular}
% \end{adjustbox}
\label{tab:ablation_detection}
\end{table*}

\begin{table*}[]
\centering
\caption{Report generation results of ablation study on the validation set of MS-CXR+MIMIC-CXR. 
"AT": abnormality token; "LE": location embedding; "Cls. Sup.": classification supervision by image-based abnormalities; "CoE": co-evolutionary strategy. 
BLEU-N denotes N-gram score of BLEU \cite{papineni2002bleu}. 
The best results are shown in bold. 
{\color{blue}*: statistically significant difference from CoE-DG for pairwise comparison with the Wilcoxon signed-rank test ($p<0.05$).}}
% \begin{adjustbox}{width=\textwidth}
\begin{tabular}{l|cccc|cccccc|c}
\hline
            & DIP-AT & DIP-LE & Cls. Sup.   & CoE & {BLEU-1} & {BLEU-2} & {BLEU-3} & {BLEU-4} & {METEOR} & {ROUGE} & AUC \\ \hline
Baseline (R2Gen\cite{chen2020generating})    & & & & & 0.3124*   & 0.1889*  & 0.1253* & 0.0892* & 0.1280* & 0.2546* & 0.63\\
Ablation-1   & \checkmark & & & & 0.3253* & 0.1924* & 0.1294* & 0.0905*& 0.1306* & 0.2611* & 0.62\\
Ablation-2   & \checkmark & \checkmark & & & 0.3297* & 0.2008* & 0.1335* & 0.0950* & 0.1325* & 0.2630* & 0.64\\
Ablation-3   & \checkmark & \checkmark & \checkmark & & 0.3330* & 0.2182* &  0.1524* & 0.1117* & 0.1344* & 0.2982* & 0.67\\ 
CoE-DG (ours)   & \checkmark & \checkmark & \checkmark & \checkmark & \textbf{0.3590} & \textbf{0.2365} & \textbf{0.1664} & \textbf{0.1229} & \textbf{0.1447} & \textbf{0.3026} & \textbf{0.69}\\\hline
\end{tabular}
% \end{adjustbox}
\label{tab:ablation_generation}
\end{table*}

\subsection{Special Validation on Normal Cases}
{\color{blue}
To assess the performance of our framework on normal cases, we randomly select 30 normal cases (about the average number of the eight abnormalities in the fully labeled test set of MS-CXR) from the test split of MS-CXR+MIMIC-CXR for a special evaluation.
As these cases are all normal, we consider predicting any abnormality in a case as a false positive; otherwise, it is a true negative. 
We use the true negative rate (TNR) for the evaluation of abnormality detection and reports' clinical efficiency.
The detection and report generation results are presented in Table \ref{tab:normal-detect} and Table \ref{tab:normal-gen}, respectively.
As we can see, our framework achieves the highest TNR regarding both abnormality detection and report generation, outperforming the second-best methods by 10\% and 3.33\%, respectively.
In addition, it is superior to R2Gen by 0.0085--0.0255 for the language-efficiency metrics.
These results validate the efficacy of the proposed weak supervision by reports in reducing false positives, and demonstrate the robustness of our framework on normal cases.
}

\subsection{Ablation Study}\label{sec:exp:ablation}
We conduct ablation studies on the validation data to investigate the efficacy of the novel building elements of our {CoE-DG} framework.
For abnormality detection, the ablated components include self-adaptive non-maximum suppression (SA-NMS), generator-guided information propagation (GIP) with feature enhancement (FE) by concatenating the detector- and generator-extracted image embeddings and pseudo label refinement (PLR) guided by generator-classified abnormalities, and co-evolutionary (CoE) strategy. 
We use the preliminary teacher-to-student pseudo label distillation as baseline (Eqn.  (\ref{eq:pld})). 
Table \ref{tab:ablation_detection} shows that introducing SA-NMS (Ablation-1) improves performance by up to 1.42\% in mAP@0.75 compared with the baseline. 
{\color{blue}An alternative strategy to incorporate the knowledge of the student into the pseudo labels for dynamic adaptation is to update the teacher with the exponential moving average (EMA) of student models (Table \ref{tab:ablation_detection}, Ablation-2).
However, introducing EMA does not bring definite performance improvement upon the baseline, and our SA-NMS consistently outperforms EMA with all three IoU thresholds.}
% \todo{An alternative strategy is to update the teacher with the Exponential Moving Average (EMA) of the student model instead of SA-NMS.
% However, we found no clear improvement with the baseline, even performing worse than the baseline in terms of mAP@0.25 and mAP@0.5 (Ablation-2).}
In Ablation-3 and Ablation-4, when combining the FE and PLR in GIP, the performance is boosted by 2.06\%--3.29\%.
Lastly, by adopting CoE, our full model achieves further improvements of up to 2.39\% (CoE-DG).

For report generation, the ablated components include detection-guided information propagation (DIP) with abnormality token (AT) and location embedding (LE), extra supervision by image-based abnormality classes (Cls. Sup.), and co-evolutionary (CoE) strategy.
The R2Gen\cite{chen2020generating} is used as the baseline.
The results are shown in Table \ref{tab:ablation_generation}.
By incorporating AT and LE, we observe improvements of 0.0082--0.0119 in language-efficacy metrics compared with the baseline (Ablation-1 and Ablation-2), indicating the effectiveness of DIP.
Moreover, when the model is additionally supervised by the multi-(pseudo-)label classification task (Ablation-3), further improvements of up to 0.0352 in ROUGE and 0.03 in AUC are achieved.
Eventually, with CoE, our full model yields the best validation results for all evaluated metrics, demonstrating the efficacy of CoE again.

\begin{figure}[t]
    \centering
    \includegraphics[width=\linewidth]{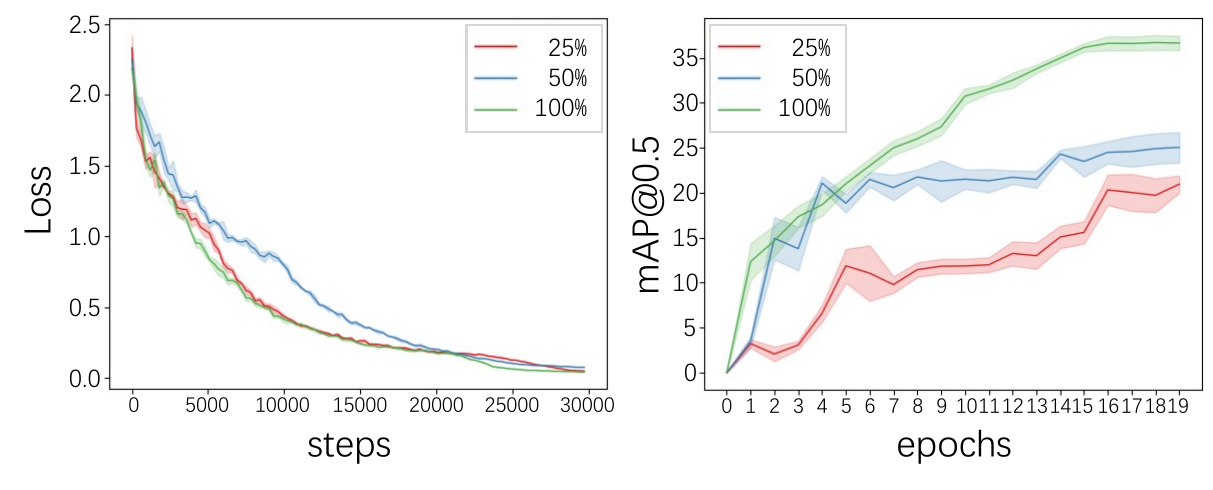}
    \caption{\color{blue}Training loss ($\mathcal{L}^I$) and validation mAP@0.50 of the detection model, using 25\%, 50\%, and 100\% of the fully labeled data of MS-CXR.
    The plot shows the mean curves (the central dark lines) of three co-evolving iterations ({iterations 0, 1, and 2 in Fig. \ref{fig:generation} left}), with corresponding spans overlaid (the light-shaded strips).
    }
    \label{fig:detection_loss}
\end{figure}

\begin{figure}[t]
    \centering
    \includegraphics[width=\linewidth]{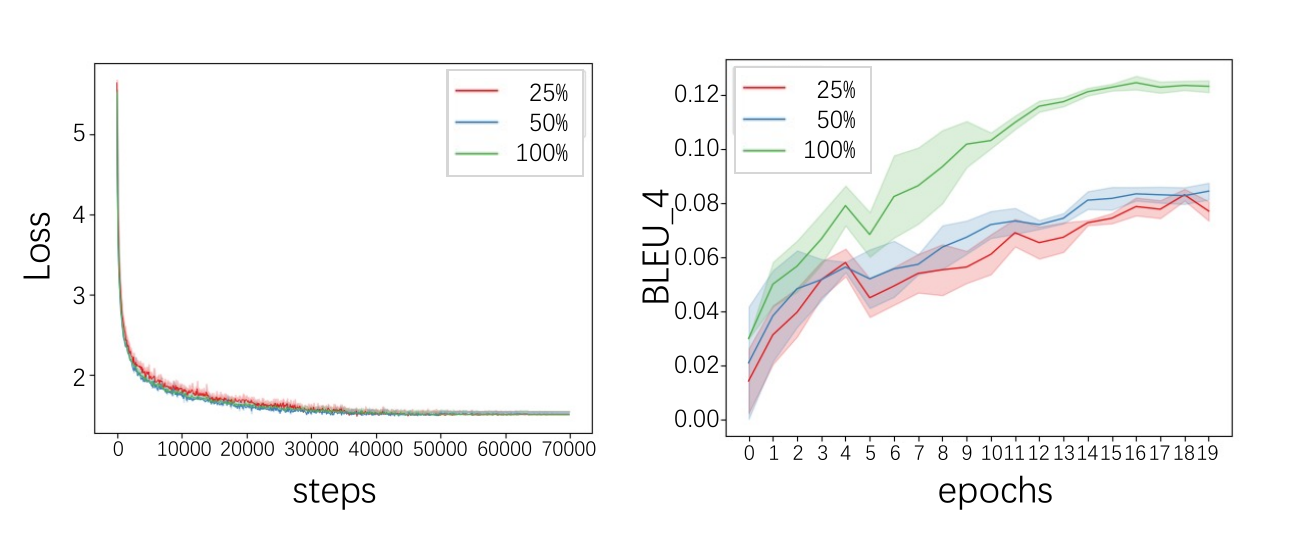}
    \caption{\color{blue}Training loss ($\mathcal{L^R})$ and validation BLUE-4 of the generation model, using 25\%, 50\%, and 100\% of the fully labeled data of MS-CXR.
    The plot shows the mean curves (the central dark lines) of three co-evolving iterations ({iterations 1, 2, and 3 in Fig. \ref{fig:generation} right}), with corresponding spans overlaid (the light-shaded strips).
    }
    \label{fig:generation_loss}
\end{figure}

\subsection{Convergence Analysis}
{\color{blue}In this work, we jointly estimate the convergence status during training based on the training loss and validation performance.
For abnormality detection, Fig. \ref{fig:detection_loss} left and right plot the curves for $\mathcal{L}^I$ (Eqn. (\ref{eq:pld})) on the training set and mAP@0.5 on the validation set of MS-CXR, respectively.
Also, we study the impact of the amount of fully labeled data, \textit{i.e.}, 25\%, 50\%, and 100\%.
As we can see, the training losses gradually decrease as training continues, and the mAPs reach plateaus in the last few epochs---for all three ratios of labeled data.
These phenomena suggest that the models are well-trained.
Fig. \ref{fig:generation_loss} shows the corresponding curves for report generation, \textit{i.e.}, $\mathcal{L}^R$ (Eqn. (\ref{eq:rgen})) on the training data and BLEU-4 on the validation data of MS-CXR+MIMIC-CXR.
The patterns are similar to those in Fig. \ref{fig:detection_loss}.
Therefore, both modules in our co-evolving framework converge well, regardless of the exact amount of labeled data.
Meanwhile, the validation performance drops slightly as the labeled data decreases, which is expected.
However, as Table \ref{tab:det_ratio} shows, our CoE-DG demonstrates superior label efficiency in comparison with the compared abnormality detection methods, with notable advantages when using 25\% and 50\% labeled data.
Lastly, it is worth noting that even 100\% of the labeled data constitutes only a tiny fraction of all training data, \textit{i.e.}, less than 1\%.
}

\subsection{Computational Costs}
{\color{blue}Table \ref{tab:cost} summarizes the numbers of parameters, giga floating-point operations (GFLOPs), GPU memory consumption, training times, and inference speeds of our framework.
In addition, the corresponding numbers for the baseline models on which our framework is built (\textit{i.e.}, RetinaNet \cite{lin2017focal} for abnormality detection and R2Gen \cite{chen2020generating} for report generation) are also presented for reference.
{As we can see, our framework only adds marginal overheads to the baselines for both tasks in model complexity, training (for one iteration), and inference.
Since our framework obtains the optimal models with three iterations, it requires roughly twice as much training time as the baseline models.
In practice, however, the inference time differences between the baseline models and ours are hardly perceivable, and the inference memory differences ($\leq0.5$GB) are largely insignificant for modern hardware.}
Therefore, from the perspective of practical use, we regard the negligible differences in inference efficiency as an acceptable trade-off for the substantial accuracy improvements of our framework.}
% In terms of the number of parameters and GFLOPs, our framework only adds marginal overheads on the baseline models for both the abnormality detection and report generation tasks.
% For training, while the training times per epoch are similar, ...

\begin{table}[]\color{blue}
\centering
\caption{Model complexity and computational costs of our CoE-DG framework.
The evaluation is carried out with NVIDIA V100 GPUs on the MS-CXR+MIMIC-CXR dataset.
Corresponding numbers for the baseline models on which our framework is built (\textit{i.e.}, RetinaNet \cite{lin2017focal} for abnormality detection, and R2Gen \cite{chen2020generating} for report generation) are also presented for reference.}
\label{tab:cost}
\begin{adjustbox}{width=\linewidth}
\setlength{\tabcolsep}{0.8mm}
\begin{tabular}{c|c|cc|cc|cc}
\hline
\multirow{2}{*}{Task} & \multirow{2}{*}{Model} & \multicolumn{2}{c|}{Model complexity} & \multicolumn{2}{c|}{Train (one iteration)} & \multicolumn{2}{c}{Inference} \\ \cline{3-8} 
                      &                        & Parameters          & GFLOPS          & Per epoch   & GPU Mem.   & Per sample    & GPU Mem.  \\ \hline
\multirow{2}{*}{Detection}             & RetinaNet              & 49.1M               & 113.6           & 0.58h        & 27.3GB       & 0.05s         & 1.5GB         \\
                      & CoE-DG                 & 55.5M               & 127.1           & 0.62h        & 30.9GB       & 0.13s         & 2.0GB         \\ \hline
\multirow{2}{*}{Generation}             & R2Gen                  & 79.8M               & 534.1           & 0.85h          & {5.8GB}        & 1.16s         & 1.6GB         \\
                      & CoE-DG                 & 81.3M               & 536.2           & 0.92h          & {10.0GB}       & 1.74s         & 1.9GB         \\ \hline
\end{tabular}
% \begin{tabular}{l|l|cc|cc|cc}
% \hline
%                             &            & Train / Epoch & \multicolumn{1}{l|}{Memory Usage} & Test/sample & \multicolumn{1}{l|}{Memory Usage} & Parameters & \multicolumn{1}{l}{GFLOPS} \\ \hline
% \multirow{2}{*}{Generation} & R2Gen      & 1h            & 5827M                             & 1.21s       & 1575M                             & 79.8M      & 534.1                      \\
%                             & R2Gen+ours & 1h            & 9970M                             & 2.04s       & 1905M                             & 81.3M      & 536.2                      \\ \hline
% \multirow{2}{*}{Detection}  & RetinaNet  & 0.5h          & 27252M                            & 0.05s       & 1543M                             & 49.1M      & 113.6                      \\
%                             & CoE-DG     & 0.6h          & 30884M                            & 0.13s       & 1997M                             & 55.5M      & 127.1                      \\ \hline
% \end{tabular}
% The dimension $M$ is set to 1024. The learning rate is set to $5\times 10 ^{-5}$ for the feature extractor and $10^{-4}$ for other parameters.
\end{adjustbox}
\end{table}

\section{Discussion}
CXR abnormality detection and report generation are two essential tasks in the clinical routine. 
Abnormality detection involves identifying abnormal regions and structures in CXR images, such as pneumonia, consolidation, and pleural effusions.
Accurate abnormality detection is crucial for disease diagnosis and treatment planning. 
Meanwhile, report generation involves describing and summarizing the detected abnormalities into a comprehensive medical report, which is essential for clinicians to make informed decisions.
Report generation requires not only accurate abnormality detection but also the ability to analyze and interpret the detected abnormalities in a meaningful way. 
Previous studies often focused on either abnormality detection or report generation. 
For example, some studies \cite{yu2022anatomy,wu2020automatic} focused on improving the accuracy of abnormality detection without considering report generation, whereas others \cite{chen2020generating,wang2023metransformer} {ignored} the rich information in the detection task while improving the generated reports.
%Similarly, report generation models\cite{chen2020generating,wang2023metransformer} ignore the rich information in the detection task. 
%In our opinion, 
Considering both tasks separately might limit their potential. 
To this end, we proposed a co-evolutionary information interaction framework that integrated both tasks for not only localizing and identifying abnormalities more accurately but also generating better reports. 
Considering both tasks together, the abnormality detection model learned from the report generation model and vice versa. 
The models shared information regarding the detected abnormalities, enabling the detection model to refine its predictions and the generation model to produce more accurate and clinically relevant reports.

\begin{figure}[t]
    \centering
    \includegraphics[width=0.8\linewidth]{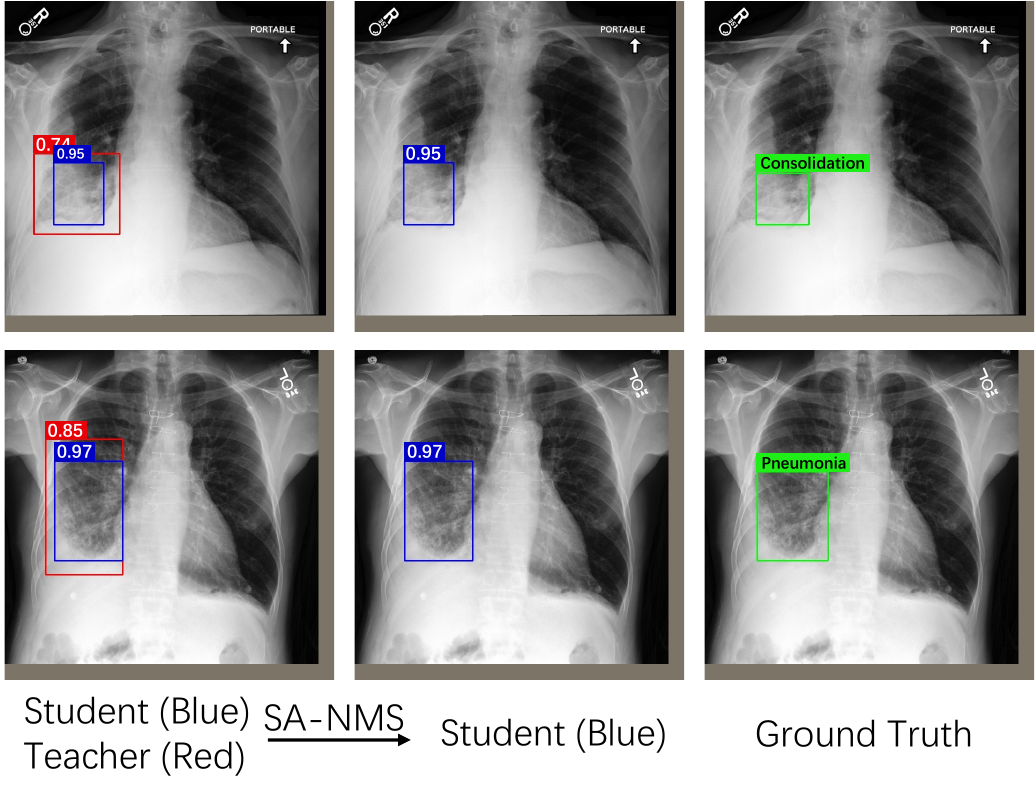}
    \caption{\color{blue}Examples demonstrating the results before and after applying the proposed self-adaptive NMS (SA-NMS).
    The prediction probabilities are overlaid on the bounding boxes (best viewed zoomed in).
    The SA-NMS keeps the more confident predictions by the student model, which are closer to the ground truth.}
    \label{fig:SA-NMS-effect}
\end{figure}

In the comparison to SOTA detection approaches (Table \ref{tab:det_res}), we compared four groups of methods: weakly supervised, supervised baseline (by fully labeled data only), semi-supervised, and unified vision-language models. 
The results showed that the weakly supervised methods were generally worse than the supervised baseline and semi-supervised methods, suggesting that the use of labels was still significant. 
However, semi-supervised methods designed for natural images become suboptimal when applied to medical images, likely due to small lesions and blurred object boundaries. 
%However, semi-supervised methods designed for natural images cannot be directly applied to medical images because the lesion areas are small in medical images, and the boundaries are unclear. 
Thus, pseudo labels generated by the teacher model might contain large noise.
%, which may result in suboptimal performance of the model. 
To filter the pseudo labels and alleviate the impact of noise, we proposed SA-NMS to adaptively combine the predictions by the teacher and student models for training, leading to improved performance (Table \ref{tab:ablation_detection}, Ablation-1).
{\color{blue}Fig. \ref{fig:SA-NMS-effect} shows two examples demonstrating the results before and after applying SA-NMS. 
As we can see, SA-NMS keeps the more confident predictions by the student model, which are closer to the ground truth. 
The visualization illustrates the benefits of SA-NMS.}
{\color{blue}Meanwhile, the alternative strategy, EMA, did not bring definite performance improvement upon the baseline (Table \ref{tab:ablation_detection}, Ablation-2).
We conjecture this was because the student model lacked sufficient detection capability at the beginning, and updating its parameters to the teacher model might harm the detection capability of the teacher. 
This, in turn, harmed the student model's training. 
In contrast, our SA-NMS rectifies the pseudo labels predicted by the teacher model with the student model’s predictions only when the latter is more confident than the former. 
In this way, the teacher and student models cooperate to gradually improve the quality and robustness of the pseudo detection labels as the training continues.}
%Concretely, we first adopt an SA-NMS that combines the pseudo labels predicted by teacher and student models, substantially improving the detection performance (Table. \ref{tab:ablation_detection}, Ablation-1).
%The results confirm that SA-NMS can reduce the influence of the fixed teacher model by continuously learning from the student model.

In addition, the semi-supervised methods did not make use of the accompanying reports.
On the contrary, despite leveraging the reports (in $D^l$) for pretraining, the unified vision-language models could not improve detection performance compared to the semi-supervised methods, likely because they did not utilize weakly labeled data for abnormality detection.
%This could be due to the failure in leveraging the potential of the weakly labeled data.
% and the lack of a specific design for medical report generation (e.g., memory-driven transformer \cite{chen2020generating}). 
Our proposed CoE-DG nicely combined both approaches with a generator-guided information propagation (GIP) module.
On the one hand, our abnormality detection model was built on the classical teacher-student distillation paradigm for semi-supervised learning.
On the other hand, a report generator was trained by the CXR reports,
%The generator was mainly supervised by reports, 
which were much more in quantity and provided different perspectives than the bounding box annotations.
Then, we refined the pseudo detection labels guided by generator-classified abnormalities.
Furthermore, we enhanced the detector-extracted image feature by appending the generator-extracted to it.
Thus, we implicitly boosted the detection model with the supervision by reports, as evidenced in Table \ref{tab:ablation_detection}, Ablation-2 and Ablation-3.
%To this end, we proposed a DIP module to optimize the detection model using the predictions of the generation model.
%Then, we refined the pseudo labels (Table. \ref{tab:ablation_detection}, Ablation-3) of the detection model by utilizing the classification results predicted by the generation model. 
%The reason behind this approach is that the class token in the generation model has learned comprehensive information about the report, including specific descriptions of normal and abnormal regions. Therefore, this information-rich class token can be utilized to supervise weakly labeled images and improve detection performance.
%Furthermore, we concatenated the refined abnormal embedding as a supplement to the input features of the detection model (Table. \ref{tab:ablation_detection}, Ablation-2). Motivated by the fact that the abnormal features extracted by the generation model, through attention mechanism training, differ from the points of interest of the target detection model, feature concatenation can enable the detection model to focus on features it has not previously considered.
The architecture of CoE-DG allowed effective leverage of all available data and supervision signals---not only the weakly labeled images but also their weak labels (\textit{i.e.}, reports) for abnormality detection.
%These approaches allowed us to effectively leverage weakly labeled data and improve the precision of pseudo labels, ultimately enhancing the performance of the detection model. 
%The performance dropped with less labeled data, as expected, but still best in comparison.

{\color{blue}Our preliminary work, CEIRD \cite{sun2023you}, focused solely on abnormality detection and could not generate text reports for CXRs.
In contrast, the more comprehensive successor, CoE-DG, can also generate reports, thus holding greater clinical significance.
Furthermore, CoE-DG achieved superior abnormality detection results to CEIRD (Table \ref{tab:det_res}).
Concretely, while CEIRD consistently outperformed other methods, CoE-DG outperformed CEIRD with all three IoU thresholds.
We attribute the improvements to the feature enhancement (FE) in the GIP module (GIP-FE) by concatenating the detector- and generator-extracted image embeddings. 
The efficacy of GIP-FE was quantitatively validated in our ablation experiments (Table \ref{tab:ablation_detection}, Ablation-1 versus Ablation-3).
This is likely because the generator-extracted image embeddings were supervised by the report generation loss, providing a different and supplementary perspective to the detector-extracted embeddings.
Therefore, not only was CoE-DG more capable than CEIRD in generating text reports, but CoE-DG was superior to CEIRD in detecting abnormalities.}

% Jinghan's original:
% This improvement can be attributed to several key factors: a) We have introduced a feature enhancement strategy that incorporates textual features as input, while CEIRD relied solely on image input. This integration of textual features provides additional context and information, thereby improving the model's performance.
% b) We utilize the classification prediction of the class token to filter pseudo-labels for the detection model. This strategy further refines the learning process and enhances the accuracy of our model.
% These enhancements underscore the advancements we have made from our previous work, leading to the superior performance of CoE-DG.

\begin{figure}[t]
\centering
\includegraphics[width=\linewidth]{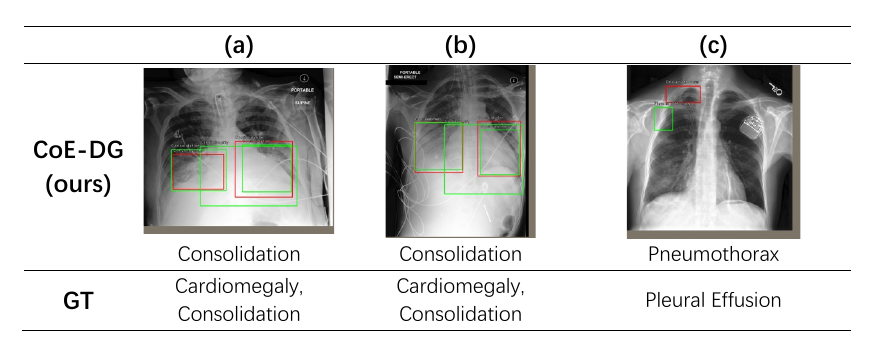}
\caption{\color{blue}Visualization of failure detection cases by our framework.
The green and red boxes are the ground truth (GT) and predictions, respectively.}
\label{fig:vis-fail}
\end{figure}

{\color{blue}Fig. \ref{fig:vis-fail} presents some cases where our CoE-DG framework failed in detection.
% mainly when there is a significant overlap between lesions (a, b) or when the lesion is blurry and small (c). In these challenging scenarios, our method may not perform as accurately as expected.
In Figs. \ref{fig:vis-fail}(a) and \ref{fig:vis-fail}(b), our framework missed the large cardiomegaly lesions that overlapped significantly with two smaller consolidation lesions.
We conjecture this was because the substantial overlap obscured the distinctive features of the cardiomegaly lesion.
% {\color{red}(Not due to NMS elimination?)}
% Discussion: Overlapping Lesions: When lesions overlap significantly, it becomes challenging for the model to discern between different abnormalities. 
% This is because the distinct features of each lesion might be obscured due to the overlap, making it difficult for the model to accurately identify and classify each abnormality.
In Fig. \ref{fig:vis-fail}(c), our framework missed the pleural effusion but falsely detected a pneumothorax, both tiny.
This might be because (1) tiny and blurry lesions were challenging to discern due to difficulty extracting effective representation, and (2) both types of lesions were under-represented in the experimental dataset.
% Small, Blurry Lesions: Small and blurry lesions present another challenge as their features are not clearly defined. The model might struggle to detect these lesions because they lack the distinct characteristics that the model has been trained to recognize.
We expect to employ more advanced baseline detection models in the future to reduce both types of failures.
}
%are fewer misses (b, c) and better localization (a).

To validate our framework's performance on report generation, we compared it with existing SOTA models and unified vision-language models (Table \ref{tab:rep_res}).
The unified models did not perform as well as the exclusive SOTA models.
This might be because the unified models were originally proposed for short phrase generation of natural images, whereas medical reports are typically lengthy and complex.
%We found that the performance of the unified model was even worse than the baseline report generation model. 
%This may be because the unified model's generation objective is limited to short phrases, while medical reports are complex. 
In contrast, exclusive report generation models \cite{chen2020generating} performed better with specialized memory modules designed to handle long texts of multiple sentences.
However, most SOTA report generation models \cite{chen2020generating,chen2022cross} focused on generating linguistically high-quality reports (\textit{e.g.}, text overlap measured by BLEU \cite{papineni2002bleu}) but paid little attention to clinical efficacy (\textit{e.g.}, measured by AUC).
%However, SOTA report generation models \cite{chen2020generating,chen2022cross} often do not pay attention to the disease category but only ensure consistency in the text. 
With a small set of bounding box annotations, our CoE-DG framework improved both language generation and clinical efficacy metrics upon existing SOTA models with detector-guided information propagation (DIP).
%In our work, we proposed a detector-guided information propagation (DIP). 
Specifically, we concatenated abnormality tokens and location embedding as input to the report generator---both extracted from the detector's output and empirically proved effective in generating better reports (Table \ref{tab:ablation_generation}, Ablation-1 and Ablation-2).
On the one hand, the detector's output directed the generator's attention to candidate abnormal regions, making the latter less prone to omitting clinically relevant findings than generating reports from whole images.
On the other hand, the location embedding provided clues for diseases of strong location preferences and for describing the abnormalities' whereabouts. 
In addition, we used the pseudo abnormality labels predicted by the detection model for additional supervision of the generator, further encouraging the latter to be aware of abnormalities in CXRs while generating reports.
The benefits were evidenced in Table \ref{tab:ablation_generation}, Ablation-3 and Table \ref{tab:rep_res} by improved language- and clinical-efficacy metrics.
% \textbf{A side benefit of introducing the additional pseudo label supervision was that the generator could also directly output abnormality categories in addition to reports for CXRs.}

%Specifically, we used the pseudo labels predicted by the detection model to supervise the generation model, ensuring that the generated reports correctly contain the diagnosed disease, and this supervision can improve the quality of generated reports (Table. \ref{tab:ablation_generation}, Ablation-3).
%In addition, by concatenating the abnormal embedding and position embedding extracted by the detection model as inputs, our approach improved upon baseline (Table. \ref{tab:ablation_generation}, Ablation-1 and Ablation-2). 
% There are several possible explanations for this result.
% CT images contain a large amount of information, and some lesions of specific categories may be tiny, so the model may directly ignore these regions. 
% Therefore, we used the detection results as inputs to make the generation model pay more attention to the abnormal area. 
% Moreover, the location of the abnormality is usually related to the category of the lesion, so we used position embeddings as part of the input to provide more information to the generation model.
% Experimental results showed that our CoE-DG effectively supplies additional knowledge for the generation process.

Moreover, the detection and generation models were bridged by our proposed co-evolutionary (CoE) strategy, such that they mutually boosted the performance of each other.
%the improved detection model can further assist in training a better generation model as  
Specifically, the detection model provided precise information about the locations and types of abnormalities to enhance the generation model's understanding of input CXRs and improve its ability to generate accurate reports.
Inversely, the generation model helped the detection model filter noisy pseudo labels and enhanced features for detection with features extracted from the perspective of report generation.
%Conversely, the generation model can provide supplementary information to the detection model, guiding it to pay more attention to critical areas and features in the CXRs. 
As shown in Fig. \ref{fig:generation}, Table \ref{tab:ablation_detection}, and Table \ref{tab:ablation_generation}, our CoE strategy led to substantial improvements in both the detection and generation performance.

{\color{blue}Lastly, our CoE-DG framework is agnostic to the exact report generation model used and can incorporate any typical model for report generation. In this work, we experimented with incorporating three baseline report generation models, \textit{i.e.}, R2Gen \cite{chen2020generating}, R2GenCMN \cite{chen2022cross}, and ORGan \cite{hou2023organ}, in our framework.
The results showed that incorporating them brings improvements upon the original models in almost all language-efficiency metrics on two datasets. 
In addition, the AUCs improve by noticeably 2\%--6\% on both datasets.
These results indicated the effectiveness of our DIP and CoE strategies in enhancing report generation—especially in terms of clinical efficiency, and the applicability of our framework.
}

% \textbf{A limitation of this work lied in the requirement for partially annotated datasets with bounding box annotations.} 
% Therefore, datasets without bounding box annotations are not applicable. In the future, we will explore leveraging self-supervised and transfer learning to fully use unlabeled data without bounding box annotations to connect detection and generation tasks.

% 1. Open-world ( our method assumes a fixed set of abnormality categories) (multi-view Chest X-ray studies or a sequence of studies)
% 2. multi-modal

{\color{blue}In our experiments, the medical large multimodal models (\textit{i.e.}~MLMMs), LLaVA-Med \cite{li2024llava}, CheXagent \cite{chen2024chexagent}, and XrayGPT \cite{thawkar2023xraygpt}, yielded generally low numbers for the language-efficiency metrics (BLEU-1 to BLEU-4, METEOR, and ROUGE) for the task of report generation.
We conjecture this was because the MLMMs' primary target, \textit{i.e.}, instruction-following / conversational systems, was misaligned with the specific task.
As a result, the MLMMs did not generate texts closely following the language pattern, structured format, or specific terminologies of radiology reports.
In addition, as LLaVA-Med was trained on general-medicine figures, captions, and corresponding in-line descriptions extracted from PubMed Central articles, it might lack enough expertise in understanding CXRs, leading to unsatisfactory clinical efficacy (low AUCs), too.
In contrast, CheXagent and XrayGPT were specialized in interpreting CXRs and achieved AUCs close to those of R2Gen and R2GenCMN (two exclusive report generation models), despite their low language-efficiency scores.
This phenomenon calls for more appropriate metrics, beyond traditional ones, for evaluating medical texts generated by modern large language models.
It is worth mentioning that our framework substantially outperformed the MLMMs in terms of both language and clinical efficiency.
These results indicated the advantages of specialist models and the effectiveness of our proposed visual detector guided report generation.
We believe it is promising to develop visually grounded MLMMs specialized in CXR report generation in the future.
}

This study had limitations. 
It only considered the subsets of abnormality categories with bounding box annotations provided by the MS-CXR {\color{blue}and PD-CXR} datasets, whereas the MIMIC-CXR dataset included 14 types of anatomical abnormalities.
%First, our work is constrained by the fixed set of eight abnormality categories, which may not be sufficient to cover all possible types and subtypes of abnormalities. 
In the future, techniques for open-set learning \cite{joseph2021towards,zheng2022towards} may be investigated and adapted to detect the categories without bounding box annotations but described in clinical reports.
%we plan to investigate methods for diagnosing abnormality categories outside the fixed set. 
We shall aim in the future to evaluate our methodology on clinical image data beyond CXR to show broader practicality.
% Another limitation is that our work only focuses on CXR images and does not consider other medical imaging modalities. In the future, we will explore the use of more imaging modalities, such as MRI, to assist in other medical tasks.

\section{Conclusion}
{\color{blue}This work presented a \textbf{co}-\textbf{e}volutionary semi-supervised abnormality \textbf{d}etection and report \textbf{g}eneration (CoE-DG) framework for simultaneous anatomical abnormality detection and report generation of chest X-rays.
% In this work, we proposed a new co-evolutionary abnormality detection and report generation (CoE-DG) framework for bridging semi-supervised anatomical abnormality detection and report generation in chest X-ray (CXR). 
% For abnormality detection, based on a preliminary teacher-student pseudo label distillation, we first presented self-adaptive NMS to mingle highly confident predictions by both the teacher and student for improved pseudo labels.
% We then proposed detector-guided information propagation (DIP) using the detector-extracted abnormality tokens and location embedding to propagate lesion information and the detected abnormalities to supervise the generator.
% Meanwhile, we further proposed a reverse, generator-guided information propagation (GIP) that used the generator-extracted image embedding to enhance image features and the generator-classified abnormalities to eliminate unmatched pseudo labels.
% In addition, we implemented a co-evolution strategy (CoE) that looped the GIP and DIP to iteratively optimize the abnormality detection and report generation models in an alternative manner.
Experimental results showed that CoE-DG outperformed up-to-date detection and report generation methods and that its designs were efficient.
}

% \newpage
\bibliographystyle{ieeenat_fullname}
\bibliography{tmi}

\begin{thebibliography}{10}
\providecommand{\url}[1]{#1}
\csname url@samestyle\endcsname
\providecommand{\newblock}{\relax}
\providecommand{\bibinfo}[2]{#2}
\providecommand{\BIBentrySTDinterwordspacing}{\spaceskip=0pt\relax}
\providecommand{\BIBentryALTinterwordstretchfactor}{4}
\providecommand{\BIBentryALTinterwordspacing}{\spaceskip=\fontdimen2\font plus
\BIBentryALTinterwordstretchfactor\fontdimen3\font minus
  \fontdimen4\font\relax}
\providecommand{\BIBforeignlanguage}[2]{{%
\expandafter\ifx\csname l@#1\endcsname\relax
\typeout{** WARNING: IEEEtran.bst: No hyphenation pattern has been}%
\typeout{** loaded for the language `#1'. Using the pattern for}%
\typeout{** the default language instead.}%
\else
\language=\csname l@#1\endcsname
\fi
#2}}
\providecommand{\BIBdecl}{\relax}
\BIBdecl

\bibitem{qin2018computer}
C.~Qin, D.~Yao, Y.~Shi, and Z.~Song, ``Computer-aided detection in chest
  radiography based on artificial intelligence: {A} survey,'' \emph{Biomed.
  Eng. Online}, vol.~17, no.~1, pp. 1--23, 2018.

\bibitem{monshi2020deep}
M.~M.~A. Monshi, J.~Poon, and V.~Chung, ``Deep learning in generating radiology
  reports: A survey,'' \emph{Artif. Intell. in Medicine}, vol. 106, p. 101878,
  2020.

\bibitem{lakhani2017deep}
P.~Lakhani and B.~Sundaram, ``Deep learning at chest radiography: {A}utomated
  classification of pulmonary tuberculosis by using convolutional neural
  networks,'' \emph{Radiology}, vol. 284, no.~2, pp. 574--582, 2017.

\bibitem{ougul2015lung}
B.~B. O{\u{g}}ul, P.~Ko{\c{s}}ucu, A.~{\"O}z{\c{c}}am, and S.~D. Kanik, ``Lung
  nodule detection in {X}-ray images: a new feature set,'' in \emph{Eur. Conf.
  IFMBE}.\hskip 1em plus 0.5em minus 0.4em\relax Springer, 2015, pp. 150--155.

\bibitem{rajpurkar2017chexnet}
P.~Rajpurkar \emph{et~al.}, ``{CheXNet}: {R}adiologist-level pneumonia
  detection on chest {X}-rays with deep learning,'' \emph{arxiv Preprint
  arxiv:1711.05225}, 2017.

\bibitem{liu2021exploring}
F.~Liu, X.~Wu, S.~Ge, W.~Fan, and Y.~Zou, ``Exploring and distilling posterior
  and prior knowledge for radiology report generation,'' in \emph{CVPR}, 2021,
  pp. 13\,753--13\,762.

\bibitem{chen2020generating}
Z.~Chen, Y.~Song, T.-H. Chang, and X.~Wan, ``Generating radiology reports via
  memory-driven transformer,'' in \emph{EMNLP}, 2020, pp. 1439--1449.

\bibitem{chen2022label}
B.~Chen \emph{et~al.}, ``Label matching semi-supervised object detection,'' in
  \emph{CVPR}, 2022, pp. 14\,381--14\,390.

\bibitem{sohn2020simple}
K.~Sohn, Z.~Zhang, C.-L. Li, H.~Zhang, C.-Y. Lee, and T.~Pfister, ``A simple
  semi-supervised learning framework for object detection,'' \emph{arxiv
  Preprint arxiv:2005.04757}, 2020.

\bibitem{xu2021end}
M.~Xu \emph{et~al.}, ``End-to-end semi-supervised object detection with soft
  teacher,'' in \emph{CVPR}, 2021, pp. 3060--3069.

\bibitem{hinton2015distilling}
G.~Hinton, O.~Vinyals, and J.~Dean, ``Distilling the knowledge in a neural
  network,'' \emph{arxiv Preprint arxiv:1503.02531}, 2015.

\bibitem{you2021aligntransformer}
D.~You, F.~Liu, S.~Ge, X.~Xie, J.~Zhang, and X.~Wu, ``{AlignTransformer}:
  {H}ierarchical alignment of visual regions and disease tags for medical
  report generation,'' in \emph{MICCAI}.\hskip 1em plus 0.5em minus 0.4em\relax
  Springer, 2021, pp. 72--82.

\bibitem{yang2021crossing}
Z.~Yang \emph{et~al.}, ``Crossing the format boundary of text and boxes:
  Towards unified vision-language modeling,'' \emph{arxiv Preprint
  arXiv:2111.12085}, 2021.

\bibitem{cho2021unifying}
J.~Cho, J.~Lei, H.~Tan, and M.~Bansal, ``Unifying vision-and-language tasks via
  text generation,'' in \emph{ICLR}.\hskip 1em plus 0.5em minus 0.4em\relax
  PMLR, 2021, pp. 1931--1942.

\bibitem{gupta2022towards}
T.~Gupta, A.~Kamath, A.~Kembhavi, and D.~Hoiem, ``Towards general purpose
  vision systems: {An} end-to-end task-agnostic vision-language architecture,''
  in \emph{CVPR}, 2022, pp. 16\,399--16\,409.

\bibitem{hu2021unit}
R.~Hu and A.~Singh, ``Uni{T}: {M}ultimodal multitask learning with a unified
  {T}ransformer,'' in \emph{CVPR}, 2021, pp. 1439--1449.

\bibitem{girshick2014rich}
R.~Girshick, J.~Donahue, T.~Darrell, and J.~Malik, ``Rich feature hierarchies
  for accurate object detection and semantic segmentation,'' in \emph{CVPR},
  2014, pp. 580--587.

\bibitem{boecking2022making}
B.~Boecking \emph{et~al.}, ``Making the most of text semantics to improve
  biomedical vision--language processing,'' in \emph{ECCV}.\hskip 1em plus
  0.5em minus 0.4em\relax Springer, 2022, pp. 1--21.

\bibitem{tam2020weakly}
L.~K. Tam, X.~Wang, E.~Turkbey, K.~Lu, Y.~Wen, and D.~Xu, ``Weakly supervised
  one-stage vision and language disease detection using large scale pneumonia
  and pneumothorax studies,'' in \emph{MICCAI}.\hskip 1em plus 0.5em minus
  0.4em\relax Springer, 2020, pp. 45--55.

\bibitem{johnsonmimic}
A.~Johnson \emph{et~al.}, ``{MIMIC-CXR-JPG}-chest radiographs with structured
  labels,'' \emph{PhysioNet}, 2019.

\bibitem{sun2023you}
J.~Sun \emph{et~al.}, ``You’ve got two teachers: Co-evolutionary image and
  report distillation for semi-supervised anatomical abnormality detection in
  chest {X}-ray,'' in \emph{MICCAI}.\hskip 1em plus 0.5em minus 0.4em\relax
  Springer, 2023, pp. 363--373.

\bibitem{redmon2016you}
J.~Redmon, S.~Divvala, R.~Girshick, and A.~Farhadi, ``You only look once:
  Unified, real-time object detection,'' in \emph{CVPR}, 2016, pp. 779--788.

\bibitem{lin2017focal}
T.-Y. Lin, P.~Goyal, R.~Girshick, K.~He, and P.~Doll{\'a}r, ``Focal loss for
  dense object detection,'' in \emph{ICCV}, 2017, pp. 2980--2988.

\bibitem{girshick2015fast}
R.~Girshick, ``Fast {R-CNN},'' in \emph{ICCV}, 2015, pp. 1440--1448.

\bibitem{he2017mask}
K.~He, G.~Gkioxari, P.~Doll{\'a}r, and R.~Girshick, ``Mask {R-CNN},'' in
  \emph{ICCV}, 2017, pp. 2961--2969.

\bibitem{Cheng2019}
P.~Cheng, ``3rd place solution for the 2018 {RSNA Pneumonia Detection
  Challenge},'' https://github.com/pmcheng/rsna-pneumonia, 2019.

\bibitem{liu2020self}
S.~Liu, Z.~Li, and J.~Sun, ``Self-{EMD}: {S}elf-supervised object detection
  without {ImageNet},'' \emph{arxiv Preprint arXiv:2011.13677}, 2020.

\bibitem{zhang2021weakly}
D.~Zhang, J.~Han, G.~Cheng, and M.-H. Yang, ``Weakly supervised object
  localization and detection: A survey,'' \emph{IEEE TPAMI}, vol.~44, no.~9,
  pp. 5866--5885, 2021.

\bibitem{huang2020comprehensive}
Z.~Huang, Y.~Zou, B.~Kumar, and D.~Huang, ``Comprehensive attention
  self-distillation for weakly-supervised object detection,'' \emph{NeurIPS},
  vol.~33, pp. 16\,797--16\,807, 2020.

\bibitem{cheng2020high}
G.~Cheng, J.~Yang, D.~Gao, L.~Guo, and J.~Han, ``High-quality proposals for
  weakly supervised object detection,'' \emph{IEEE TIP}, vol.~29, pp.
  5794--5804, 2020.

\bibitem{zhou2016learning}
B.~Zhou, A.~Khosla, A.~Lapedriza, A.~Oliva, and A.~Torralba, ``Learning deep
  features for discriminative localization,'' in \emph{CVPR}, 2016, pp.
  2921--2929.

\bibitem{bhalodia2021improving}
R.~Bhalodia \emph{et~al.}, ``Improving pneumonia localization via
  cross-attention on medical images and reports,'' in \emph{MICCAI}.\hskip 1em
  plus 0.5em minus 0.4em\relax Springer, 2021, pp. 571--581.

\bibitem{yu2022anatomy}
K.~Yu, S.~Ghosh, Z.~Liu, C.~Deible, and K.~Batmanghelich, ``Anatomy-guided
  weakly-supervised abnormality localization in chest {X}-rays,'' in
  \emph{MICCAI}.\hskip 1em plus 0.5em minus 0.4em\relax Springer, 2022, pp.
  658--668.

\bibitem{bearman2016s}
A.~Bearman, O.~Russakovsky, V.~Ferrari, and L.~Fei-Fei, ``What's the point:
  Semantic segmentation with point supervision,'' in \emph{{ECCV}}.\hskip 1em
  plus 0.5em minus 0.4em\relax Springer, 2016, pp. 549--565.

\bibitem{ji2022point}
H.~Ji \emph{et~al.}, ``Point beyond class: A benchmark for weakly
  semi-supervised abnormality localization in chest {X}-rays,'' in
  \emph{MICCAI}.\hskip 1em plus 0.5em minus 0.4em\relax Springer, 2022, pp.
  249--260.

\bibitem{jeong2019consistency}
J.~Jeong, S.~Lee, J.~Kim, and N.~Kwak, ``Consistency-based semi-supervised
  learning for object detection,'' \emph{NeurIPS}, vol.~32, pp.
  10\,759--10\,768, 2019.

\bibitem{tang2021proposal}
P.~Tang, C.~Ramaiah, Y.~Wang, R.~Xu, and C.~Xiong, ``Proposal learning for
  semi-supervised object detection,'' in \emph{WACV}, 2021, pp. 2291--2301.

\bibitem{jing2017automatic}
B.~Jing, P.~Xie, and E.~Xing, ``On the automatic generation of medical imaging
  reports,'' \emph{arxiv Preprint arXiv:1711.08195}, 2017.

\bibitem{xue2018multimodal}
Y.~Xue \emph{et~al.}, ``Multimodal recurrent model with attention for automated
  radiology report generation,'' in \emph{MICCAI}.\hskip 1em plus 0.5em minus
  0.4em\relax Springer, 2018, pp. 457--466.

\bibitem{yuan2019automatic}
J.~Yuan, H.~Liao, R.~Luo, and J.~Luo, ``Automatic radiology report generation
  based on multi-view image fusion and medical concept enrichment,'' in
  \emph{MICCAI}.\hskip 1em plus 0.5em minus 0.4em\relax Springer, 2019, pp.
  721--729.

\bibitem{jin2024promptmrg}
H.~Jin, H.~Che, Y.~Lin, and H.~Chen, ``{PromptMRG}: Diagnosis-driven prompts
  for medical report generation,'' in \emph{AAAI}, vol.~38, no.~3, 2024, pp.
  2607--2615.

\bibitem{radford2021learning}
A.~Radford \emph{et~al.}, ``Learning transferable visual models from natural
  language supervision,'' in \emph{ICML}.\hskip 1em plus 0.5em minus
  0.4em\relax PMLR, 2021, pp. 8748--8763.

\bibitem{zhang2023biomedclip}
S.~Zhang \emph{et~al.}, ``{BiomedCLIP}: a multimodal biomedical foundation
  model pretrained from fifteen million scientific image-text pairs,''
  \emph{arXiv Preprint arXiv:2303.00915}, 2023.

\bibitem{achiam2023gpt}
J.~Achiam \emph{et~al.}, ``{GPT}-4 technical report,'' \emph{arXiv Preprint
  arXiv:2303.08774}, 2023.

\bibitem{touvron2023llama}
H.~Touvron \emph{et~al.}, ``{LLaMA}: Open and efficient foundation language
  models,'' \emph{arXiv Preprint arXiv:2302.13971}, 2023.

\bibitem{gu2021domain}
Y.~Gu \emph{et~al.}, ``Domain-specific language model pretraining for
  biomedical natural language processing,'' \emph{ACM Trans. Comput.
  Healthcare}, vol.~3, no.~1, pp. 1--23, 2021.

\bibitem{lee2020biobert}
J.~Lee \emph{et~al.}, ``{BioBERT}: a pre-trained biomedical language
  representation model for biomedical text mining,'' \emph{Bioinf.}, vol.~36,
  no.~4, pp. 1234--1240, 2020.

\bibitem{alayrac2022flamingo}
J.-B. Alayrac \emph{et~al.}, ``Flamingo: a visual language model for few-shot
  learning,'' \emph{NeurIPS}, vol.~35, pp. 23\,716--23\,736, 2022.

\bibitem{wang2023cogvlm}
W.~Wang \emph{et~al.}, ``{CogVLM}: Visual expert for pretrained language
  models,'' \emph{arXiv Preprint arXiv:2311.03079}, 2023.

\bibitem{liu2024visual}
H.~Liu, C.~Li, Q.~Wu, and Y.~J. Lee, ``Visual instruction tuning,''
  \emph{NeurIPS}, vol.~36, 2024.

\bibitem{li2024llava}
C.~Li \emph{et~al.}, ``{LLaVA-Med}: Training a large language-and-vision
  assistant for biomedicine in one day,'' \emph{NeurIPS}, vol.~36, pp.
  28\,541--28\,564, 2024.

\bibitem{thawkar2023xraygpt}
O.~Thawkar \emph{et~al.}, ``{XRayGPT}: Chest radiographs summarization using
  medical vision-language models,'' \emph{arXiv Preprint arXiv:2306.07971},
  2023.

\bibitem{chen2024chexagent}
Z.~Chen \emph{et~al.}, ``{CheXagent}: Towards a foundation model for chest
  x-ray interpretation,'' in \emph{AAAI}, 2024.

\bibitem{dosovitskiy2020image}
A.~Dosovitskiy \emph{et~al.}, ``An image is worth 16x16 words: {T}ransformers
  for image recognition at scale,'' in \emph{ICLR}, 2020.

\bibitem{chen2022cross}
Z.~Chen, Y.~Shen, Y.~Song, and X.~Wan, ``Cross-modal memory networks for
  radiology report generation,'' \emph{arxiv Preprint arXiv:2204.13258}, 2022.

\bibitem{furlanello2018born}
T.~Furlanello, Z.~Lipton, M.~Tschannen, L.~Itti, and A.~Anandkumar, ``Born
  again neural networks,'' in \emph{ICLR}, 2018, pp. 1607--1616.

\bibitem{johnson2019mimic}
A.~E. Johnson \emph{et~al.}, ``{MIMIC-CXR-JPG}, a large publicly available
  database of labeled chest radiographs,'' \emph{arxiv Preprint
  arxiv:1901.07042}, 2019.

\bibitem{everingham2009pascal}
M.~Everingham, L.~Van~Gool, C.~K. Williams, J.~Winn, and A.~Zisserman, ``The
  {PASCAL} visual object classes {(VOC)} challenge,'' \emph{Int. J. Comput.
  Vis.}, vol.~88, pp. 303--308, 2009.

\bibitem{papineni2002bleu}
K.~Papineni, S.~Roukos, T.~Ward, and W.-J. Zhu, ``{BLEU}: {A} method for
  automatic evaluation of machine translation,'' in \emph{Proc. 40th Annu.
  Meeting Assoc. for Comput. Linguistics}, 2002, pp. 311--318.

\bibitem{banerjee2005meteor}
S.~Banerjee and A.~Lavie, ``{METEOR}: {A}n automatic metric for {MT} evaluation
  with improved correlation with human judgments,'' in \emph{Proc. ACL Workshop
  on Intrinsic and Extrinsic Eval. Measures for Mach. Transl. and/or
  Summarization}, 2005, pp. 65--72.

\bibitem{lin2004rouge}
C.-Y. Lin, ``{ROUGE}: {A} package for automatic evaluation of summaries,'' in
  \emph{Text Summarization Branches Out}, 2004, pp. 74--81.

\bibitem{smit2020chexbert}
A.~Smit, S.~Jain, P.~Rajpurkar, A.~Pareek, A.~Y. Ng, and M.~P. Lungren,
  ``{CheXbert}: {C}ombining automatic labelers and expert annotations for
  accurate radiology report labeling using {BERT},'' \emph{arxiv Preprint
  arXiv:2004.09167}, 2020.

\bibitem{paszke2019pytorch}
A.~Paszke \emph{et~al.}, ``{PyTorch}: An imperative style, high-performance
  deep learning library,'' \emph{NeurIPS}, vol.~32, pp. 8026--8037, 2019.

\bibitem{he2016deep}
K.~He, X.~Zhang, S.~Ren, and J.~Sun, ``Deep residual learning for image
  recognition,'' in \emph{CVPR}, 2016, pp. 770--778.

\bibitem{krizhevsky2012imagenet}
A.~Krizhevsky, I.~Sutskever, and G.~E. Hinton, ``{ImageNet} classification with
  deep convolutional neural networks,'' \emph{Communications of the ACM},
  vol.~60, no.~6, pp. 84--90, 2017.

\bibitem{lin2017feature}
T.-Y. Lin, P.~Doll{\'a}r, R.~Girshick, K.~He, B.~Hariharan, and S.~Belongie,
  ``Feature pyramid networks for object detection,'' in \emph{CVPR}, 2017, pp.
  2117--2125.

\bibitem{zhang2022glipv2}
H.~Zhang \emph{et~al.}, ``{GLIPv2}: {U}nifying localization and vision-language
  understanding,'' \emph{NeurIPS}, vol.~35, pp. 36\,067--36\,080, 2022.

\bibitem{hou2023organ}
W.~Hou, K.~Xu, Y.~Cheng, W.~Li, and J.~Liu, ``{ORGAN}: Observation-guided
  radiology report generation via tree reasoning,'' in \emph{AMACL (Volume 1:
  Long Papers)}, 2023, pp. 8108--8122.

\bibitem{sun2023eva}
Q.~Sun, Y.~Fang, L.~Wu, X.~Wang, and Y.~Cao, ``{EVA-CLIP}: Improved training
  techniques for {CLIP} at scale,'' \emph{arXiv Preprint arXiv:2303.15389},
  2023.

\bibitem{wang2022medclip}
Z.~Wang, Z.~Wu, D.~Agarwal, and J.~Sun, ``{MedCLIP}: Contrastive learning from
  unpaired medical images and text,'' in \emph{EMNLP}, 2022, pp. 3876--3887.

\bibitem{wu2020automatic}
J.~Wu \emph{et~al.}, ``Automatic bounding box annotation of chest {X}-ray data
  for localization of abnormalities,'' in \emph{ISBI}, 2020, pp. 799--803.

\bibitem{wang2023metransformer}
Z.~Wang, L.~Liu, L.~Wang, and L.~Zhou, ``{METransformer}: {R}adiology report
  generation by {T}ransformer with multiple learnable expert tokens,'' in
  \emph{CVPR}, 2023, pp. 11\,558--11\,567.

\bibitem{joseph2021towards}
K.~Joseph, S.~Khan, F.~S. Khan, and V.~N. Balasubramanian, ``Towards open world
  object detection,'' in \emph{CVPR}, 2021, pp. 5830--5840.

\bibitem{zheng2022towards}
J.~Zheng, W.~Li, J.~Hong, L.~Petersson, and N.~Barnes, ``Towards open-set
  object detection and discovery,'' in \emph{CVPR}, 2022, pp. 3961--3970.

\end{thebibliography}

\end{document}